# B.E. PROJECT ON
# An Investigation into Neuromorphic ICs using Memristor-CMOS Hybrid Circuits

Submitted By:

Shikhar Makhija
155EC15

Udit Kumar Agarwal
188EC15

Varun Tripathi
196EC15

Under the Guidance of:

Dr. Kunwar Singh

A Project in partial fulfillment requirement for the award of
B.E. in

Electronics & Communication Engineering

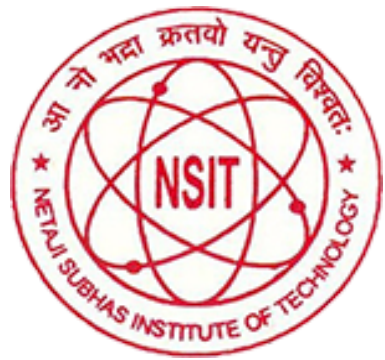

**Department of Electronics & Communication Engineering, (NSIT)**

**Now upgraded to NETAJI SUBHAS UNIVERSITY OF TECHNOLOGY**
NEW DELHI - 110078
2019

# CERTIFICATE

This is to certify that the report entitled **An Investigation into Neuromorphic ICs using Memristor-CMOS Hybrid Circuits** being submitted by Shikhar Makhija, Udit Kumar Agarwal and Varun Tripathi to the Division of Electronics and Communication Engineering, NSIT, for the award of bachelor's degree of engineering, is the record of the bonafide work carried out by them under our supervision and guidance. The results contained in this report have not been submitted either in part or in full to any other university or institute for the award of any degree or diploma.

**Supervisors**

## Dr. Kunwar Singh

ECE Department
NSIT, New Delhi, India
(now upgraded to NSUT)

# ACKNOWLEDGEMENT

We would like to express our deepest appreciation to the Department of ECE for providing a conducive environment for completing our project. We would like to thank our final year project guide, Dr. Kunwar Singh, for his constant support and guidance throughout the project and for his encouragement throughout the project.

# PLAGIARISM REPORT

**BTP**

ORIGINALITY REPORT

| %12 | %5 | %10 | % |
|---|---|---|---|
| SIMILARITY INDEX | INTERNET SOURCES | PUBLICATIONS | STUDENT PAPERS |

MATCHED SOURCE

| 3 | shodhganga.inflibnet.ac.in<br>Internet Source | %1 |
|---|---|---|

1%

★ shodhganga.inflibnet.ac.in
Internet Source

| EXCLUDE QUOTES | OFF | EXCLUDE MATCHES | OFF |
|---|---|---|---|
| EXCLUDE BIBLIOGRAPHY | ON | | |

# LIST OF ABBREVIATIONS

CMOS: Complementary Metal Oxide Semiconductor

MOS: Metal Oxide Semiconductor

MOSFET: Metal Oxide Field Effect Transistor

NMOS: n-type Metal Oxide Semiconductor

PMOS: p-type Metal Oxide Semiconductor

VLSI: Very Large Scale Integrated Circuits

HP: Hewlett Packard

TEAM: ThrEshold Adaptive Memristor Model

STDP: Spike Timing Delay Plasticity

OTA: Operational Transconductance Amplifier

ReLU: Rectified Linear Unit

CNN: Convolutional Neural Network

ANN: Artificial Neural Network

AF: Activation Function

# ABSTRACT

Memristors are passive two-terminal devices which behave similar to variable resistors. The memristance of a memristor depends on the amount of charge flowing through it and when current stops flowing through it, it remembers the state. Thus, memristors are extremely suited for implementation of memory units. Memristors find great application in neuromorphic circuits as it is possible to couple memory and processing, compared to traditional Von-Neumann digital architectures where memory and processing are separate. Neural networks have a layered structure where information passes from one layer to another and each of these layers have the possibility of a high degree of parallelism. CMOS-Memristor based neural network accelerators provide a method of speeding up neural networks by making use of this parallelism and analog computation.

In this project we have conducted an initial investigation into the current state of the art implementation of memristor based programming circuits. Various memristor programming circuits and basic neuromorphic circuits have been simulated.

The next phase of our project revolved around designing basic building blocks which can be used to design neural networks. A memristor bridge based synaptic weighting block, a operational transconductor based summing block were initially designed. We then designed activation function blocks which are used to introduce controlled non-linearity. Blocks for a basic rectified linear unit and a novel implementation for tan-hyperbolic function have been proposed. An artificial neural network has been designed using these blocks to validate and test their performance.

We have also used these fundamental blocks to design basic layers of Convolutional Neural Networks. Convolutional Neural Networks are heavily used in image processing applications. The core convolutional block has been designed and it has been used as an image processing kernel to test its performance.

# TABLE OF CONTENTS

# LIST OF FIGURES

# LIST OF TABLES

# 1. Introduction to Memristors and Neuromorphic Computing

## 1.1. Memristor

**M**emristor is the theorised 'missing' circuit element that completes the clique of conventional passive elements, namely Resistor (R), Inductor (L) and Capacitor (C). It was first envisioned and theorised by circuit theorist Leon Chua in 1971[1]. Interestingly, in 2008, a team at HP Labs proposed a method in journal '*Nature*' [2], which involved the usage of Titanium Oxide in realisation of characteristics that were proposed by Chua around 30 years earlier, leading to an upheaval in scientific communities worldwide. This new circuit element paves the possibility of multiple avenues in domains such as neuromorphic computing, reconfigurable RF antenna, Non-volatile memory and Signal Processing.

Neuromorphic computational systems are systems which mimic natural neural structures and processes using very large scale integrated analog circuits. Neuromorphic computing encapsulates not only analog circuits but digital, mixed-signal VLSI systems as well as software models inspired by the biological process such as neural networks and deep neural networks. The main allure of neuromorphic circuits in the current world is that traditional computational structures suffer from an increasing gap between the computational speed and the memory bandwidth which slows them down, as well as high power requirements to run these networks. Analog neuromorphic circuits offer lower power requirements, and memristor based technology provides a way to have coupled memory and processing.

## 1.2. Characteristics of Memristor Models

The mathematical modelling of a unit such as a memristors is based on the four quadratic analysis of relationships between various static parameters, given in the figure below:

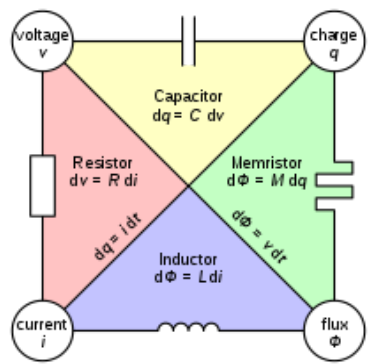

Fig 1: Mathematical Relationship between various state variables

As, one can observe there are broadly four static variables, to be considered. Two of them are independent variables while the other two are dependent on other with the help of derivative/integration mathematical operation. As, all processes in nature are usually symmetric or congruent, a logical extrapolation can be made about the missing bridge between charge and flux. The linking element is called as memristance, which is given as,

$$M = d\phi/dq \qquad\qquad -(1)$$

This can be further extrapolated, as current and voltage are easier to correlate in electronic circuits, as:

$$v = M(w)i \qquad\qquad -(2)$$

$$dw/dt = i \text{ or } w = q \qquad\qquad -(3)$$

Thus, memristance clearly depends on the amount of charge passed through it till the observation point.

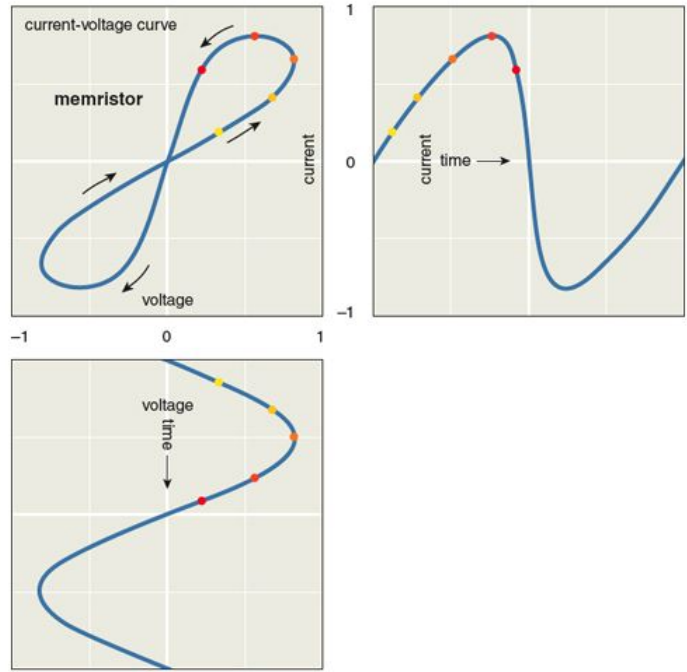

Fig 2: Memristor V-I relationship graph

## 1.3. Mathematical Models of Memristors
### 1.3.1. HP Memristor

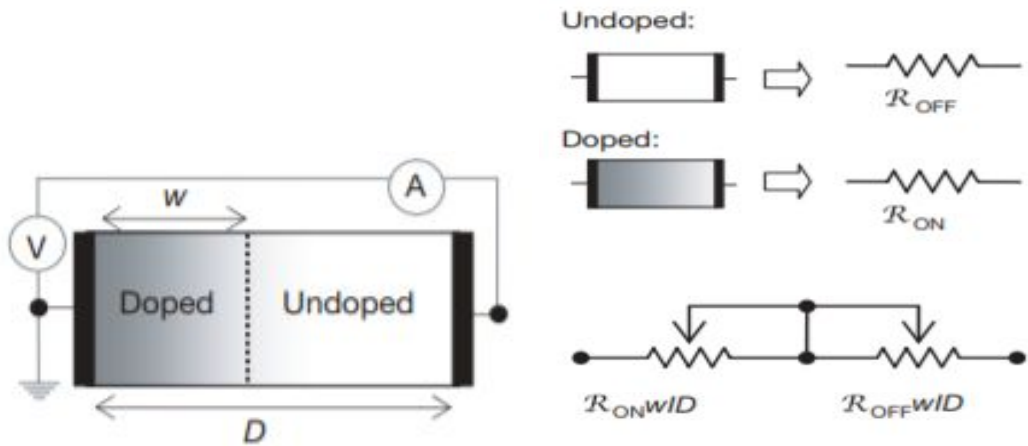

Fig 3: Coupled Variable Resistor Model

This is the memristor model proposed by HP[2*], it is based on Titanium dioxide cross points.

The equations controlling the value of memristance is given by the following equations:

$$v(t) = (R_{ON}(w(t)/D) + R_{OFF}(1 - w(t)/D))i(t) \qquad -(4)$$

$$\text{and where, } dw(t)/dt = u_v(R_{ON}/D)i(t) \qquad -(5)$$

The state variable w(t) denotes the width of doped region and the rate of change of the width of the undoped region depends on the current passing through. The value of the $R_{ON} << R_{OFF}$ , thus as more current passes through the memristance the value of the resistance offered at a time say t, decreases.

### 1.3.2. Memristor Models

A detailed analysis of these models are given below[3]:

### 1. Linear Ion Drift Model

In [2],HP developed a memristor in 2008 which exhibited a pinched hysteresis curve between current and voltage. They fabricated a titanium oxide based memristor in which an oxygen deficient layer existed. The memristance depended on the width of the oxygen deficient layer which was the state variable. [2] also performed basic modelling of the device characteristics. The basic equations proposed by Strukov et al. characterised the device based on linear ion drift where the vacancies could travel across the device, however this model is not practical as in reality as the vacancies move towards the boundary of the device they exhibit nonlinear characteristics. The original HP model assumes that the vacancies can drift across the complete device which would result in it becoming completely doped or undoped, which it does not.

### 2. Nonlinear Ion Drift Model

The non-linear ion drift model [4] takes on the some limitations in the memristor model on the account of the electrodynamics. Moreover, studies have shown the characteristics are non-linear and linear ion drift model aren't accurate enough. This model has a nonlinear relation between the state derivative and the applied voltage. This model follows asymmetric switching behavior, where during the ON state it follows the tunneling part (sinh part), while during the OFF state it follows the PN junction characteristics (exponential part).

### 3. Simons Tunnel Barrier Model

This model treats memristor as a resistor in series with an electron tunnel barrier. This model assumes non-linear and asymmetric switching behavior due to an exponential dependence of movement of the ionized dopants. This model suffers from some implicit limitations such as complexity, relationship between voltage and current is implicit and it isn't a generic model.

## 4.Threshold Adaptive Memristor Model

This model [5] is made to cater to the needs of creating a model that emulates the above model with limited error but offers simplification and computational efficiency. The model follows that there is no change in the state variable below a threshold and that there is a polynomial dependence between the memristor current and internal derivative of the state variable. TEAM supports two current-voltage models, a model which provides linear relation between memristance and the state variable and a model which allows it to fit the tunnel barrier model, where the meristance is exponentially related to the state variable.

TEAM is an effective model because as shown in [5], it in improves the simulation runtime by 47% and has a deviation of only 0.2% from other models.

Table 1 : Mathematical Relationships for memristor models (derived from [3])

| Model | Current–voltage relation | State variable derivative |
|---|---|---|
| Linear ion drift | $v(t) = \left(R_{on}\frac{w(t)}{D} + R_{off}(1 - \frac{w(t)}{D})\right)i(t)$ | $\frac{dw(t)}{dt} = \frac{\mu_v R_{on}}{D}i(t)$ |
| Nonlinear ion drift | $i(t) = w^n(t)\beta sinh(\alpha v(t)) + \chi[exp(\gamma v(t)) - 1]$ | $\frac{dw(t)}{dt} = av^m(t)f(w)$ |
| Simmons tunneling barrier | $i(t) = \tilde{A}(x, v_g)\phi_1(v_g, x) \times exp(-B(v_g, x) \cdot \phi_1^{0.5}(v_g, x)) - \tilde{A}(x, v_g)(\phi_1(v_g, x) + e|v_g|) \times exp(B(v_g, x)(\phi_1(v_g, x) + ev_g)^{0.5})$ $v_g = v - i(t)R_s$ | $\frac{dx(t)}{dt} = \begin{cases} c_{off}sinh(\frac{i}{i_{off}})exp[-exp(\frac{x-a_{off}}{w_c} - \frac{|i|}{b}) - \frac{x}{w_c}] \ i > 0 \\ c_{on}sinh(\frac{i}{i_{on}})exp[-exp(\frac{x-a_{on}}{w_c} - \frac{|i|}{b}) - \frac{x}{w_c}] \ i < 0 \end{cases}$ |
| TEAM | $v(t) = [R_{on} + \frac{R_{OFF} - R_{ON}}{x_{off} - x_{on}}(x - x_{on})].i(t)$ or $v(t) = R_{ON}.exp(\frac{\lambda}{x_{off} - x_{on}}(x - x_{on})).i(t)$ | $\frac{dx(t)}{dt} = \begin{cases} k_{off}(\frac{i(t)}{i_{off}} - 1)^{a_{off}}.f_{off}(x) & 0 < i_{off} < i \\ 0 & i_{on} < i < i_{off} \\ k_{on}(\frac{i(t)}{i_{on}} - 1)^{a_{on}}.f_{on}(x) & i < i_{on} < 0 \end{cases}$ |

A brief comparison of the various models existing in literature is shown in the following figure:

Table 2: Comparison of different memristor models (derived from [3])

| Model | Linear ion drift | Nonlinear ion drift | Simmons tunneling barrier | TEAM |
|---|---|---|---|---|
| State variable | $0 \leq w \leq D$ | $0 \leq x \leq 1$ | $a_{eff} \leq x \leq a_{on}$ | $x_{on} \leq x \leq x_{off}$ |
| Control mechanism | Current | Voltage | Current | Current |
| I–V relation | Explicit | Explicit | Ambiguous | Explicit |
| Memristance relation | Explicit | Ambiguous | Ambiguous | Explicit |
| Generic | No | No | No | Yes |
| Accuracy | Lowest | Low accuracy | Highest | Sufficient |
| Threshold exists | No | No | Yes | Yes |

## 1.3.3. Window Functions

The state variable used in memristors usually have a certain range of operation. Window functions are used to add non-linearities at the boundary regions of operation so that the state variable is confined to its operating range. For example, in the linear ion drift model, the state variable can vary between (0,D). Without a window function

f(w), the state variable can possibly go out of bounds, which would imply a completely doped or undoped semiconductor.

Properties of a window functions:

**1. Boundary conditions** – whether the boundary conditions at the top or the bottom electrode of device are taken into considerations or not.

**2. Boundary lock** – whether a window function is able account for boundary lock problem.

**3. Linearity of drift** – whether the drift at boundaries non-linear or not.

**4. Linkage between linear and non-linear model** – whether at low voltages non-linear model performs same as linear model both with same window functions.

**5. Scalability** – whether it provides full scalability or not

**6. Flexibility** – whether it provides control parameters for fine tuning several properties.

A brief compendium of the window functions offered is shown in the following table:

Table 3: Window functions

| | Joglekar | Biolek | Prodromakis | Piecewise | TEAM |
|---|---|---|---|---|---|
| $f(x)$ | $1-(2x-1)^{2p}$ | $1-(x-stp(-i))^{2p}$ | $j\left(1-\left[(x-0.5)^2+0.75\right]^p\right)$ | $\begin{cases}\left(1+\left(\frac{x-0.5}{a}\right)^{2b}\right)^{-1} & x_o \le x \le 1-x_o \\ kx(1-x) & otherwise\end{cases}$ | $exp\left[-exp\left(\frac{|x-x_{on,off}|}{w_c}\right)\right]$ |
| Symmetry | Yes | Yes | Yes | Yes | Not necessarily |
| Resolve boundary conditions | No | Discontinuities | Practically yes | Practically yes | Practically yes |
| Impose nonlinear drift | Partially | Partially | Partially | Partially | Yes |
| Scalable factor $f_{max} \le 1$ | No | No | Yes | Yes | No |
| Fits memristor model | Linear/ nonlinear ion drift/ TEAM | Linear/ nonlinear ion drift/ TEAM | Linear/ nonlinear ion drift/ TEAM | Linear/ nonlinear ion drift/ TEAM | TEAM for Simmons tunneling barrier fitting |
| | 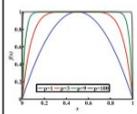 | 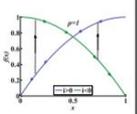 | 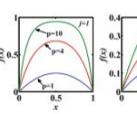 | 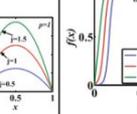 | 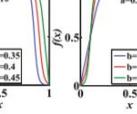 |

### 1.3.4. Neuromorphic Computer Architecture using Memristor

Neuromorphic computational systems are systems which mimic natural neural structures and processes using very large scale integrated analog circuits. Different modelling of the variable features of neuron has been proposed and developed using memristor by researchers. The main advantage that the memristor caters is its ability to feature a non-volatile memory with easy of programmability. Moreover, floating point values are also possible to achieve, offering an edge over the conventional von-Neumann architecture [6]. Due to the ease of programmability of the memristor multiple modelling have been proposed in literature modelling dendrites, Neural nets, synaptic weighting, STDP etc.

A hardware implementation allows for faster and more energy efficient implementations of artificial neural networks. [7] surveys various FPGA based implementations of neural networks. Although these implementations outperform their software counterparts a major issue faced by them is that the synaptic weights are fixed during program time and any modification made requires reprogramming of the FPGA.

Memristors offer a highly integrable solution to this problem. Memristors couple both processing and memory at one location. Memristors can also be programmed independently and don't face the drawbacks a FPGA implementation would.

[8] proposes an implementation of a simple ANN which solves linearly and nonlinearly separable Threshold Logic Unit problems using memristors as synaptic weights in the neural network.

### 1.3.4.1. Neural Network Fundamentals

Artificial Neural Networks are loosely inspired by the actual processes taking place in the nervous system. The basic element of the network is known as a 'neuron' or a node. It is the centre where processing of information takes place. Synapses interlink one neuron to another and have a weight associated with them. There is a non-linear activation function at each node with both limits the output range as well as introduces non linearity. An important aspect of Neural Networks is their layered structure where the output of one layer serves as the input of another layer. A single layer neural network is known as the Perceptron. Adaline has the structure of a perceptron and is followed by a hard-limiting activation function.

A good way of thing about Neural Networks is a "Composite function" which takes some inputs and gives and output depending on various parameters. These parameters can be updated or 'trained' for a specific application.

Neural networks are made up of:
- Neurons
- Weights/Parameters
- Biases

Neurons are the building blocks of a NN, it consists of biases and weights and performs some computation on the input to give an output. Further, Neuron uses activation functions to limit their outputs within a range.

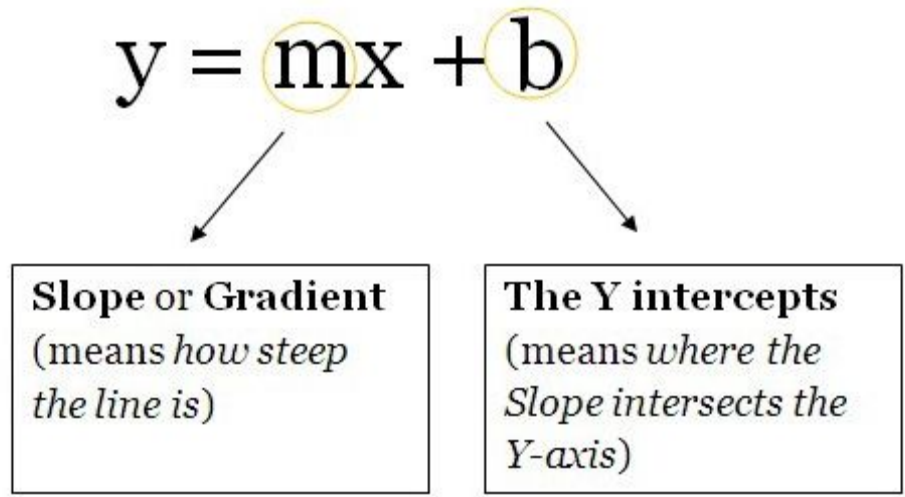

Fig. 4 : Equation of a simple neuron with variable weights and biases
{Source: **https://tinyurl.com/ya7a4bz5**}

Weights refer to interconnection weights between neurons and biases refers to additional unit that is added to a neuron's output.

A neural network can be trained by using three methods: Supervised learning, Supervised learning using critics and unsupervised learning. Supervised learning includes Back-propagation and RTRL and usually gradient descent is used to minimize the error function. Unsupervised learning follows Hebbian rule. Gradient descent is further fo 2 types: Changing the epoch and Continuous weight change.

Further there can be 2 types of neurons: Stochastic and deterministic. Deterministic neutron can be of single order or even of multiple order. Neurons can be further used in 2 topologies: Unconstrained sparse and Fully connected symmetric. Activation function in synapse can be of two types: Discrete and Continuous. Continuous activation function can further be of 3 types: pulse frequency neuron, Distributed neuron and hyperbolic tangent neuron.

Neural networks used for neuromorphic applications are of two types:

Spiking Neural networks like Spike Timing Delay Plasticity(STDP)

1. Here, Synapse plasticity depends on latency between spikes from previous and current neuron.
2. Weight of synapse increase when lag between two neurons decreases and vice versa
3. Like, Hebbian learning network which strengthens the connection between neutron whose activities are causally related.
4. STDP exploits the threshold effects observed in switching characteristics of several types of memristors

5. It's more closer to the actual working of brain

Artificial neural networks like feedforward, back propagation etc.

1. Here, Synapse plasticity depends on the weights assigned to synapse network.
2. Since, learning in ANN is iterative, it can adjust for mismatch in memristive synaptic elements.

### 1.3.4.2. Deep Neural Network Fundamentals

Supervised machine learning involves the use of a labelled set of training data to train a model which can than be used to make predictions about new data. Traditional algorithms include linear classification where the training data is used to find the coefficients of a set of linear equations which can then be used to make predictions. A simple example would be a linear classifier which classifies a small image as shown below.

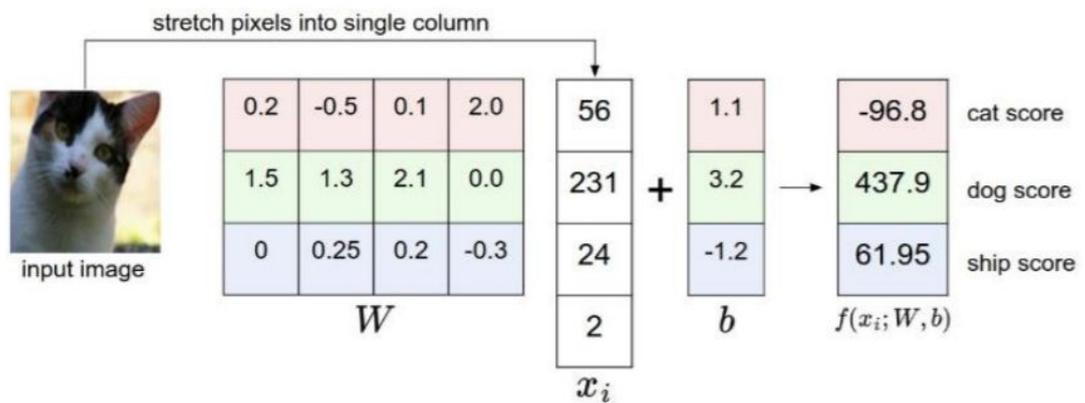

Fig 5 : Linear Classification Example {Source: CS231n Stanford 2017}

The weight matrix 'W' and bias matrix 'b' is found using the training set with algorithms which specify a loss associated with the current model. Algorithms such as gradient descent can be used to determine the optimal values of these coefficients. Neural networks are loosely inspired by the biological neuron, however in simple terms they are just linear classifiers followed by a nonlinear function which are then cascaded with each other.

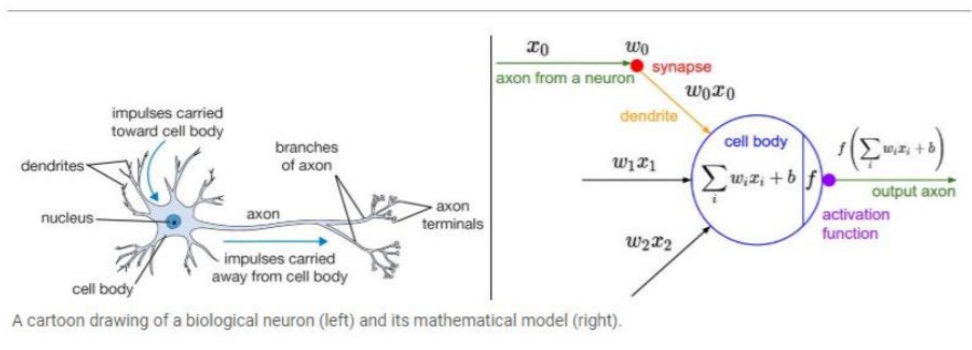

Fig 6 : Biological neuron analogy {Source: CS231n Stanford 2017}

Deep neural networks have multiple layers performing an information processing task with a number of parameters to be adjusted in each layer. A deep network layer typically consists of certain mathematical operations such as convolution, rectification linear unit (ReLU) operations, down sampling operations, concatenations, elementwise additions, batch normalization and fully connected matrix multiplications.

Convolutional neural networks are critical in modern day image processing tasks. If a neuron was attached to each pixel of an image it would take a prohibitive amount of processing time due to large sizes of images. Instead of this in convolutional neural networks a filter with weights to be trained is convolved spatially with the image to produce an output which can then serve as the input to the next layer. The concept of layers is what makes deep networks so attractive. It allows for different layer combinations which produce drastically different outputs.

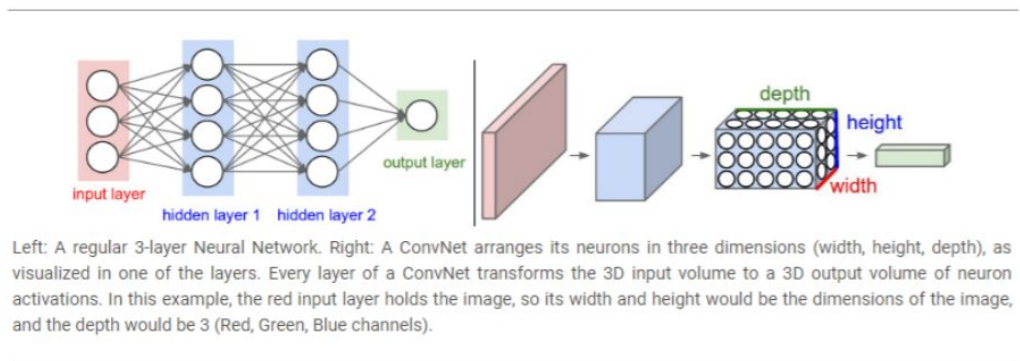

Left: A regular 3-layer Neural Network. Right: A ConvNet arranges its neurons in three dimensions (width, height, depth), as visualized in one of the layers. Every layer of a ConvNet transforms the 3D input volume to a 3D output volume of neuron activations. In this example, the red input layer holds the image, so its width and height would be the dimensions of the image, and the depth would be 3 (Red, Green, Blue channels).

Fig 7 : NN vs CNN {Source: CS231n Stanford 2017}

### 1.3.4.3. Network Learning Algorithms

Network training algorithms are the algorithms which determine the synaptic weights that need to be set. These algorithms determine these weights based on the problem to be solved and the structure of the artificial neural network being employed for the problem. Back propagation is one of the most popular training algorithms used. It is used together with gradient descent to determine the optimum weights. The error between the target response and the current response of the network is calculated and this error is propagated backwards through the entire network to set the weights. This algorithm requires a large amount of computational effort; however, it is suitable for almost every kind of network and is thus widely used.

[9] Proposes the Madaline Rule – II which is used to train multi-layer ADALINE networks which contain hard limiting functions (signum function) which is non-differentiable. The MR-II rule uses the principle of minimum disturbance to train the network. When an input is fed into the network and it responds correctly to the training values, no weight adaptation is done, otherwise, it randomly changes the weights at the input of the node that has the least confidence in its output. After the

weights are changed, the network is tested to see if the change reduced the classification error rate. If this is the case, the change is kept, otherwise, the weight is reset to the original, then another set of weights is changed. This is done according to the figure shown below. The weight change will be increased by a factor known as the growth factor if the network doesn't converge after a set number of iterations.

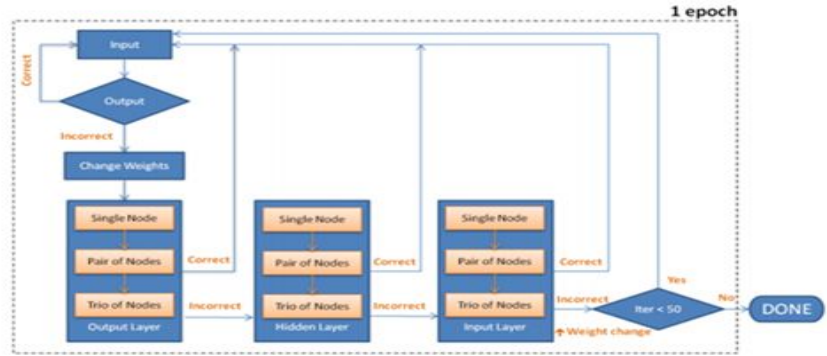

Fig.8 : MR-II Rule

## 1.4. Literature Review

### 1.4.1. Memristor and Theoretical Modelling

Leon Chua proposed in 1971 the fourth missing element [1] and the same was implemented by HP in 2008 [2]. Multiple models have been forth over the years focusing on obtaining different memristive characteristics like pinched hysteresis loop. Linear Ion Drift Model [2] was initially modelled by HP. Later, other different models like "Non-Linear Ion Drift Memristor Model" [4], "Simmon Tunnel Barrier Model" [10] and "Team Model" [5] were presented.

Each of these models has its own plus and minus. In Linear model, nonlinearity is less and assumes two conditions that are (i) uniform electric field and (ii) the average mobility of ions [11]. Further it predicts an inverse relationship between switching time($T_o$) and applied voltage($V_o$), stimuli is voltage here. However, actual experiments shows a logarithmic inverse relationship.

$$\text{Predicted Relationship: } V_0 \text{ α } 1/T_o \qquad -(6)$$
$$\text{Actual Relationship: } V_0 \text{ α } 1/T_o \qquad -(7)$$

To mitigate this problem, nonlinear memristor models were proposed. Further, window functions such as "Jogelker" [12], "Biolek" [13], and "Prodromakis" [14] play a vital role in linear model as well as non-linear model. It is used to block the state variable from getting out of the bounds [0,D] and also to add more non-linearity near the bounds. The derivative of this state variable is multiplied by a window junction in order to reduce it to 0 when the state variable is at the bounds.

### 1.4.2. Memristor for biological neuron

Memristors are mostly strong candidates for physical realization of a synapse. However, it can also be used to model biologically inspired neuron. Transient current response of memristor is exploited for the same. [15] shows that memristor gives a spike-like current response for step input and it can be used in voltage-excited neuromorphic systems. It can further be demonstrated that the change in the spike's width depends upon change in initial resistance. [16] then compared several characteristics of memristors transient response with biological systems. On-Off switching ratio between biological system is shown to be 20 times more than that of a memristive device. Further, the concept of using logarithmic amplifiers is introduced to reduce the above discrepancy. Pickett's model of memristor was used to model switching characteristics of a memristor. [17] demonstrates the use of MMOST - Memristor MOS technology for positional detector. Further, an extensive study of symmetric and asymmetric STDP was carried out and Inhibition of Return(IOR) neuromorphic algorithm was proposed. Another algorithm, kth order Winner-Take-All was proposed. [18] introduced the concept of Long-term depression , Long-term potentiation and habituation for the ways in which memory is stored or made in all living beings. Further, Presence of LTP within memristor was demonstrated. Chua further draw resemblance between Hodgkin-Huxley's sodium potassium channel model and corresponding memristor model. Anomalies in original Hodgkin-Huxley model were solved by using memristor. [19] proposed memristive models for higher order STDP and SRDP learning algorithms. Four different types of STDP's were shown and TSTDP(Triple STDP) was implemented with memristive synapse. Depression and Potentiation rates were introduced in memristive based post synaptic spikes. [20] proposed a memristor based neuron for SRDP. Switching rate of neuron was altered to change synapse's sensitivity.

### 1.4.3. VLSI Implementation of Analog Neural Networks

Multiple implementation of the CNN and Artificial Neural Network has been proposed in the past based on CMOS[21]-[23]. However, a large amount of circuit is required for implementation of it on the chip. As this goal is extremely daunting not many full architectures have been proposed in the past. For the synaptic weighing and analog multiplication, CMOS based multipliers have been commonly used in the past. But, programmable multipliers with weights have not been proposed, much in the past. Also, nonlinearity in synaptic multiplications between input and weight is also a problem of the conventional circuit.

[24, [25] and [26] presented different multiplier circuits. [24] presented a approach limited to spiking neural network, while [25] presented memristor based

self-organized network with the possibility of negative weights, Kim et al. [26] presented a generic neural network multiplier implementation.

[27] aimed at developing a full-scale single board neural network. This basically means that the stimulus generating analog circuit and sensing circuit is on the same chip. Moreover, dual modes exists for the neural network node, namely: 1. Dynamic Mode: For training purposes. 2. Permanent Mode: Retaining the charged weights. This paper gives an in detail description of the interior circuitry of the neural network. Moreover, the results of the resultant circuit for 2-XOR-6 and two-spiral problems have also been enlisted. [28] develops from ground the logic of the most basic type of an artificial neural network hardware implementation. It starts with the details about CMOS circuit and extends it to multiplier circuits which leads to the pathway for a basic neural network analog circuit. This can prove to be a good starting point to develop a VLSI implementation of the Neural Networks. [29] can also be used as a starting point for learning CMOS VLSI chip implementation of neural networks. The researchers in this paper developed a chip containing 3K Transistors arranged into a matrix of 8x4 synapses fully connected to 4 neurons. It was proposed that this rudimentary model, is then used to develop more sophisticated structures. Error analysis due to the analog anomilities is also presented at the end of the paper in brief. [30] discusses about Hopfield ADC which is based on the feed forward neural network and compares it with the Flash ADC and the SAR ADC on the basis of the propagation delay. Moreover, it compares with another neural network, named as the asymmetrical neural network which is an advancement to the Hopfield ADC in terms of elimination of the spurious energy level condensation problem. [31] proposes a VLSI neural network in which the learning process requires minimal additional hardware i.e. clock, inputs and a target value. The readjustment of the weights during the learning phase is done using approximation method such as Medeline Rule III and summed weight neuron perturbation. Resultant arrangement was tested on a small 64 synapse, 8 electron model on $4900um^2$ of chip testing with the 4-parity problem. A number of varied points related to electronic implementation of neural networks were proposed. In analogy with the human brain , an analog implementation of neural networks will be pursued using simple, small, possibly non-ideal building blocks; neurons and synapses. [32] presents an analog neural network as an application for the Support Vector Machine learning based on a partially dual formulation of the quadratic programming problem.

### 1.4.4. Memristor Programming

[33] proposes an antisymmetric series memristor combination in which the weights can be programmed linearly. [34] builds upon this architecture using them to create memristor bridge synapses which use a combination of the above antisymmetric series memristor combinations to create weights which are stable at zero. They maintain the linear programming property and also allow for easy interconnection of

weights. This in contrast to the traditional memristor programming models as in [35] where a memristors in a crossbar are programmed to a desired weight. The weights in this case are non-zero and it does not follow the linear programming regime.

## 2. An Investigation into State of the Art Memristor Circuits

### 2.1. Memristor Programming circuit

A memristor can be programmed to a fixed value precisely by giving a specific number of voltage pulses. One such implementation of the above logic is shown below:

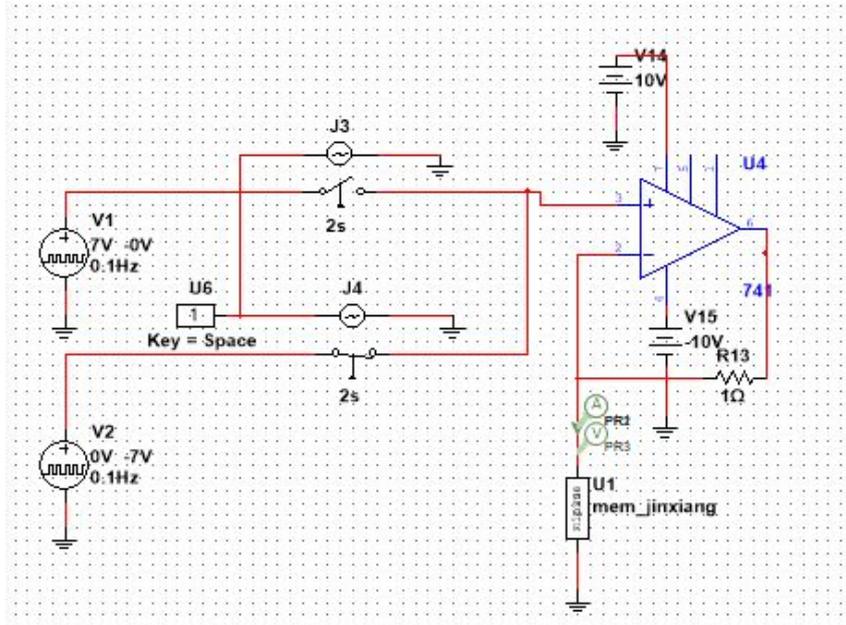

Fig. 9a : Memristor programming circuit

For this demonstration, we have used a Non-Linear memristor model with the JinXiang window function. Transient analysis of the above circuit is shown below, where the memristance is changed from one extreme i.e. Roff to another extreme i.e. Ron. The disadvantage with the given configuration was that the programming of the memristors was non-linear.

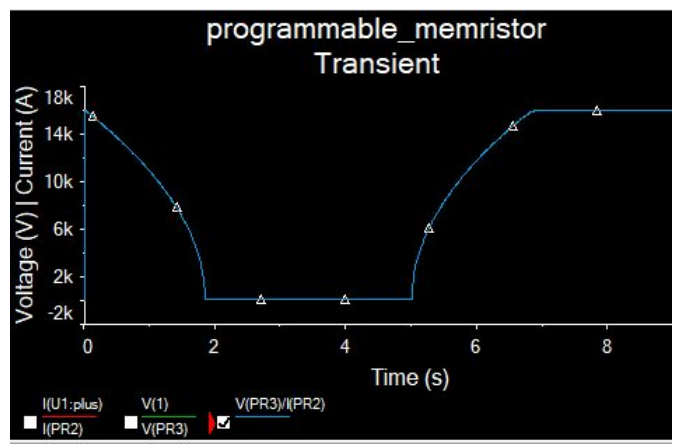

Fig 9b : Memristance curve while programming

## 2.2. CMOS Memristor Dendrite Threshold Circuit

[37] designs dendritic circuits of spike and saturation types. Dendrites form the interface between synapses in a biological neuron. These circuits can be used to build XOR circuits as well as intensity detection circuits which are formed by a combination of these basic dendrites. Circuits build in [37] use memristors, CMOS gates and zener diodes.

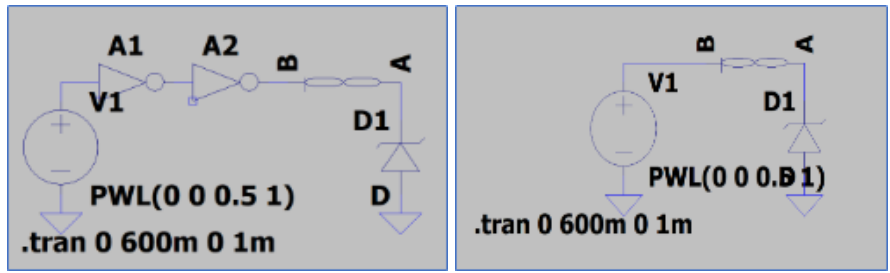

(a)  Dendrite Spike Type Cell          (b) Dendrite Saturation Type Cell

Fig 10 : Dendrite simulation

The outputs of the circuit were observed as:

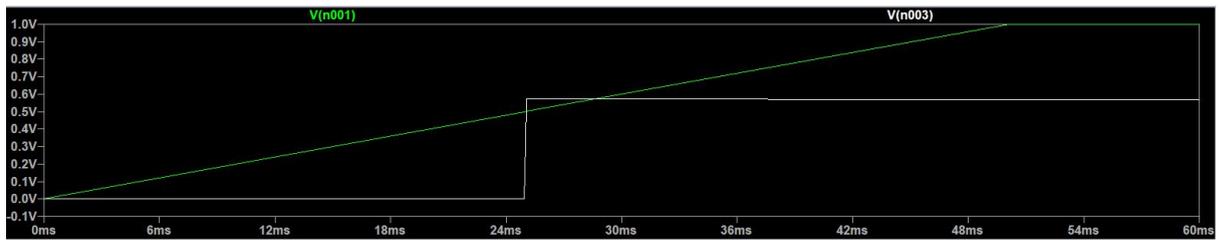

(a)

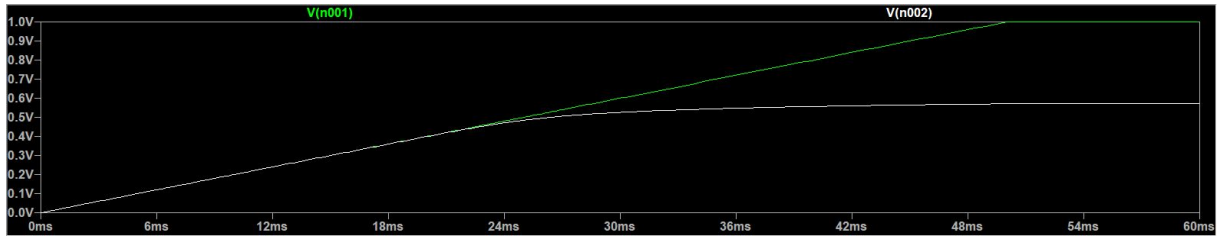

(b)

Fig 11: Different Dendrite Models (a) Step (b) Threshold dendrite

Spike and saturation dendrites and a XOR circuit built using them was simulated. As our main objective was to design a kernel based subsystem, the edge detection method using dendrite based multiple threshold circuit was explored in depth.

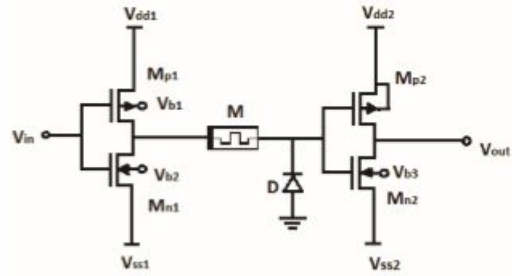

Fig 12: Multiple Threshold circuit

The not gate using the CMOS logic used are variable threshold type not gates. A not gate and the memristor-zener in the above configuration forms the inverse dendritic logic. The practical response is that it offers image segmentation and can be used in applications such as edge detection. By tuning the threshold of the above circuit different colour regions can be processed. This circuit is useful for edge detection and image segmentation where pixels are classified into groups.

The input image pixel vector was used as the basis for creating variable voltage inputs which were fed to the above circuitry. The inverse logic of the above was used to create the output images. Some of the sample outputs we simulated are shown in the following figures.

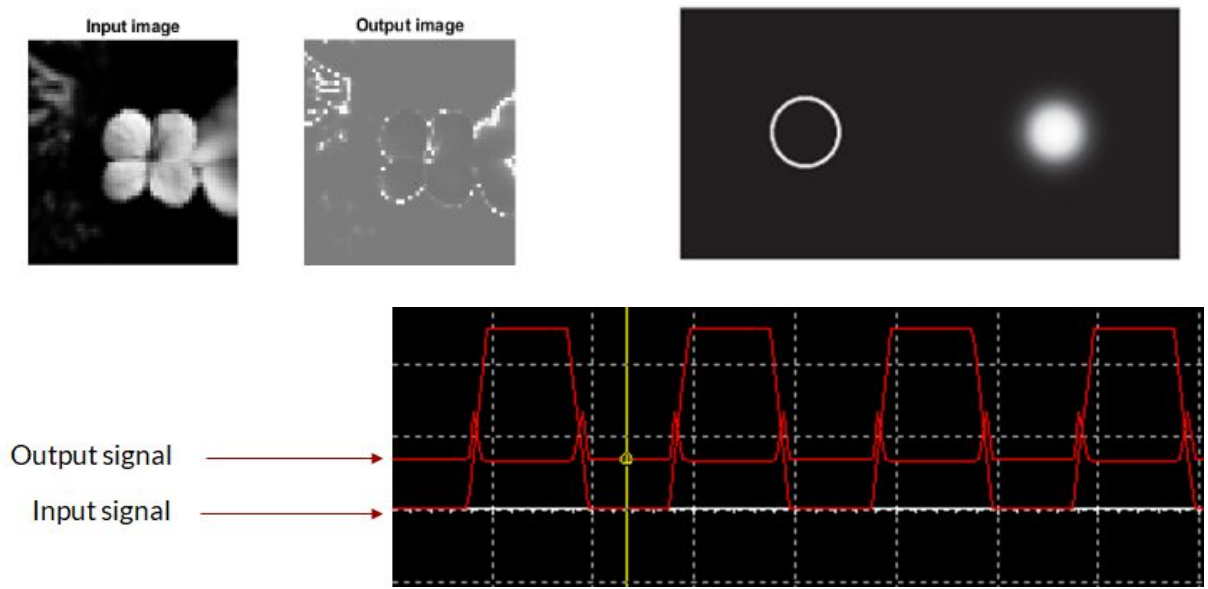

Fig 13: Output of Simulations

The approach thus, developed offered us a great insight on a basic neuromorphic circuit for the application of edge detection in images. This circuitry doesn't directly offer us any advantage in the kernel making. But, certain amolerations can be made to it, providing a great lead:

Use of different zener to control the saturation level. However, a reverse control voltage can also be used for controlling the breakdown of the zener and increase the domain of the proposed.

## 2.3. Memristor Bridge Circuit for Neural Synaptic Weighting

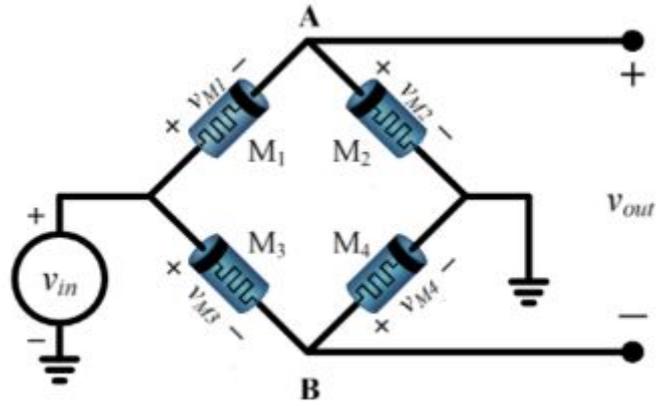

Fig 14: Memristor Bridge Circuit

The above memristor bridge circuit [38] was the basis of the proposed research approach. The basic voltage equations are given below:

$$V_A = M_2 v_{in}/(M_1 + M_2) \qquad -(8)$$

$$V_B = M_3 v_{in}/(M_3 + M_4) \qquad -(9)$$

$$V_{OUT} = V_A - V_B = (M_2/(M_1 + M_2) - M_3/(M_3 + M_4))v_{in} = \varphi v_{in} \quad -(10)$$

$M_1$, $M_2$, $M_3$ and $M_4$ are the memristance at time instance t.

$$\varphi = \begin{cases} +, & \dfrac{M_2}{M_1} > \dfrac{M_4}{M_3} \\ -, & \dfrac{M_2}{M_1} < \dfrac{M_4}{M_3} \end{cases}$$

$$-(11)$$

Interestingly, the equivalent constant $\varphi$ is linear in nature, thus offering linear synaptic control.

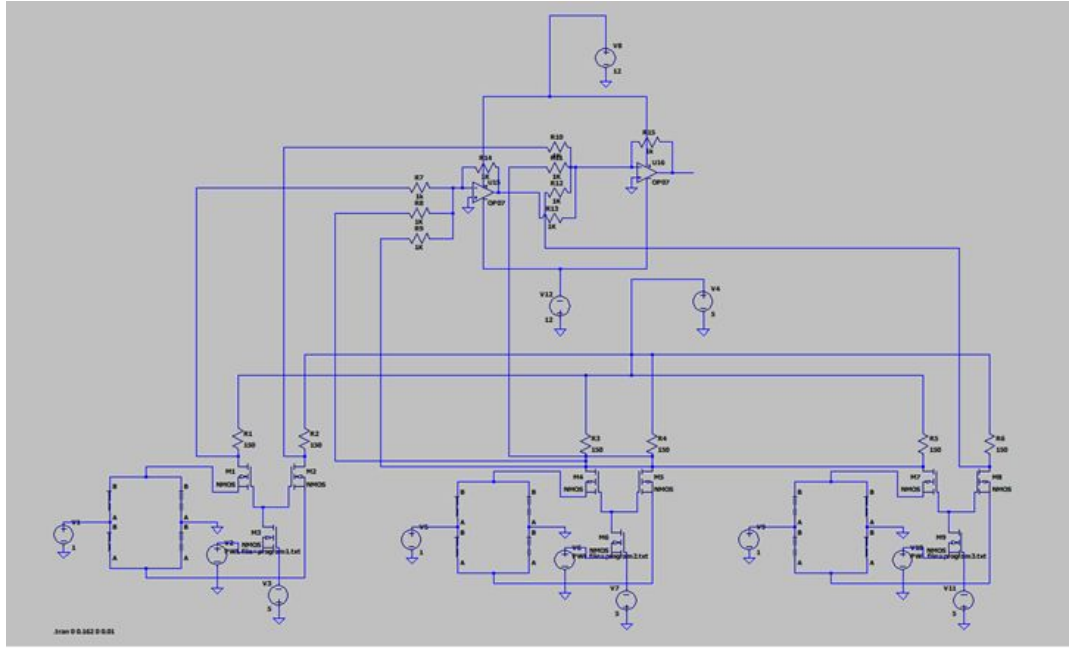

Fig 15a: Proposed Synaptic Weighting System

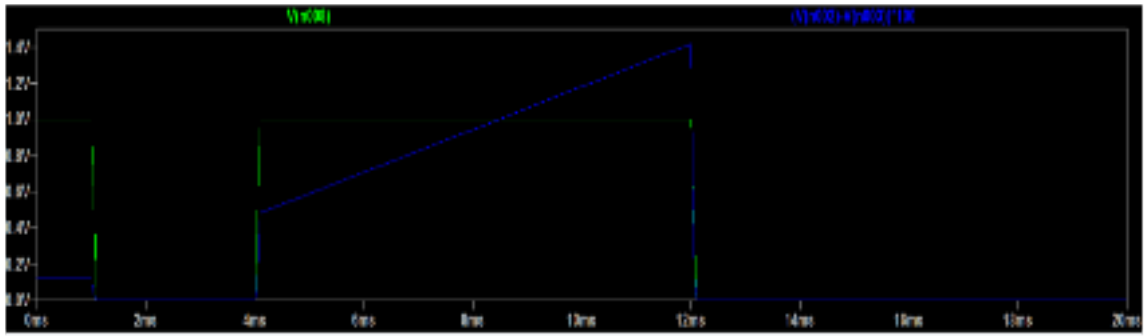

Fig 15b: Programming memristance on a bridge network

The programming mechanism of this circuit is based on input-time logic. For smaller time periods the resultant shift in memristance is small, while on the other hand the movement in memristance for larger change in input signal voltage is large.

Following, the logic above we designed a neural synaptic weighting circuit using opamp which is given below:

The circuit utilizes voltage summing and resultant voltage differencing over the following stage. This circuit offers a basis for programming the synaptic weights to get a programmable feed forward stage of NN.

The output of this circuit over the programming and weighting stage is shown in the following figure.

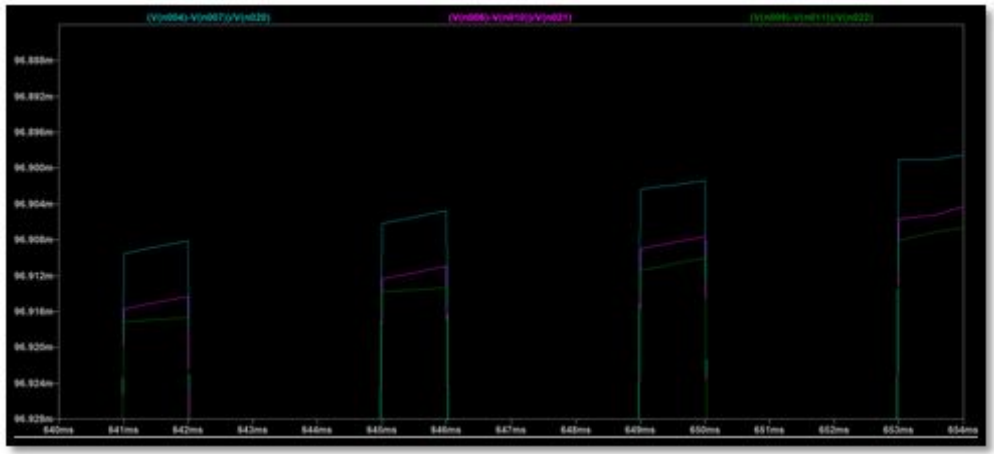

Fig 16: Synaptic weights at the time of calculating the outputs

The reference bias values are summarized in the following table.

Table 4: Pulse Programming timing for the synaptic weights

| Synaptic Weight | Time Period for Programming(s) |
|---|---|
| 1 | 0.639 |
| 2 | 0.319 |
| 3 | 0.159 |

The programming is done on this circuit using a PWL file written in python. These are taken as inputs, as a text file.

**2.4.Implementation of linearly separable TLUs:**

Linearly separable TLUs (Threshold Logic Units) require only a single line to separate their regions. They can easily be implemented using a single neuron with two inputs and a bias term. There is a large number of possible weight combinations which satisfy the relation of a given TLU. To implement negative synaptic weights the following circuit has been proposed.

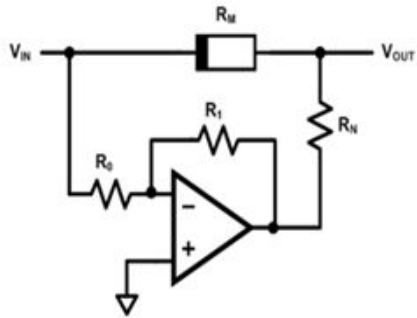 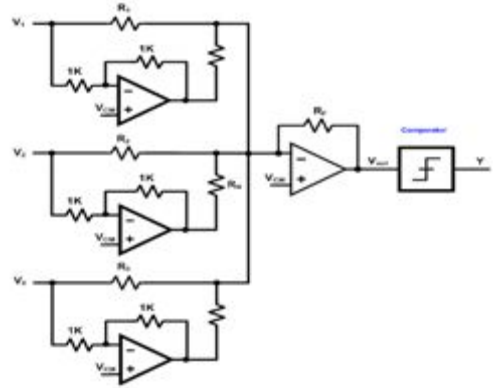

Fig.17. Negative synaptic weight          Fig.18. Fixed Resistor Model

The overall structure of the proposed single node Adaline circuit is shown below:

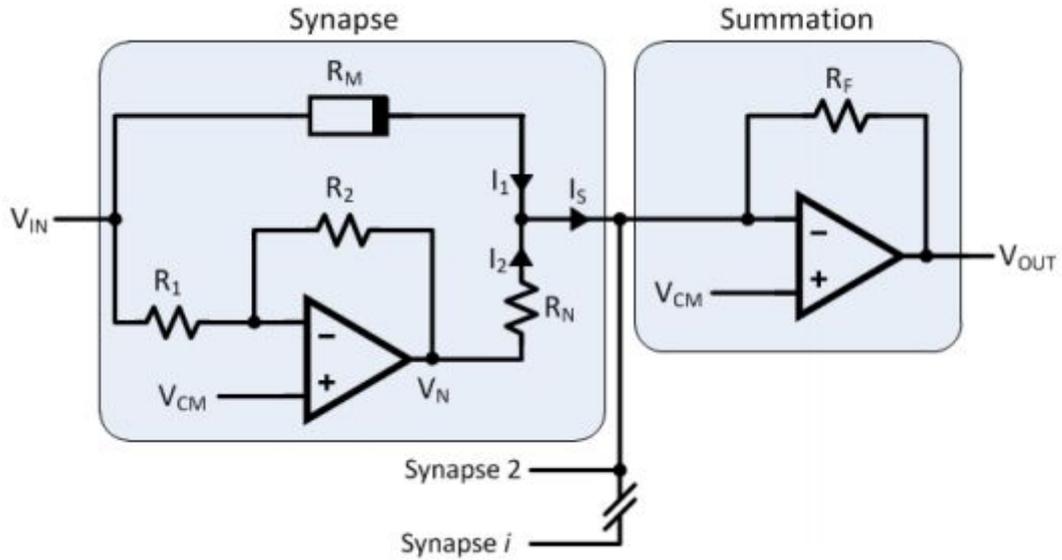

Fig.19. Single Node Structure

The equations used to map the weights determined to the given structure are shown below:

$$R_N = R_{M_{HIGH}} - (R_{M_{HIGH}} G_{HIGH})(R_{M_{HIGH}} - R_{M_{LOW}})/(R_{M_{HIGH}} G_{HIGH} - R_{M_{LOW}}) \quad -(12)$$

$$R_F = R_N(R_{M_{HIGH}} G_{HIGH} - R_{M_{LOW}} H_{LOW})/(R_{M_{HIGH}} - R_{M_{LOW}}) \quad -(13)$$

where, $R_{M_{HIGH}}$ = Highest Memristance Value

$R_{M_{LOW}}$ = Lowest Memristance Value

$G_{HIGH}$, $G_{LOW}$ = Maximum and Minimum weight values found

The resistance values obtained using these equations are as follows:

Table 5 : Resistance values

|  | $R_1$ | $R_2$ | $R_0$ |
|---|---|---|---|
| NAND | 1.33 KΩ | 1.17 KΩ | 3.88 KΩ |
| NOR | 1.33 KΩ | 1.17 KΩ | 1.33 KΩ |
| AND | 2.81 KΩ | 4.81 KΩ | 1.33 KΩ |
| OR | 2.81 KΩ | 4.81 KΩ | 3.88 KΩ |

The single layer Adaline circuit was then simulated in LTSPICE:

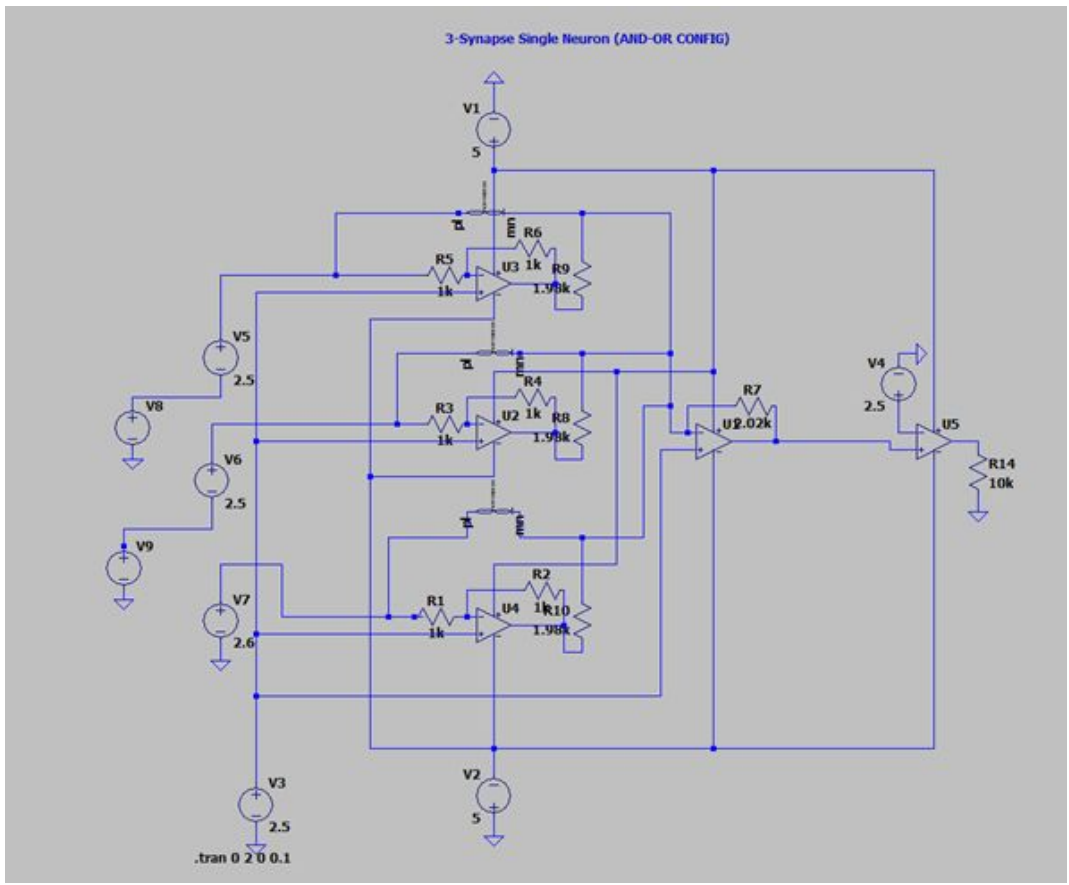

Fig 20. Single Neuron Adaline Structure

The Memristance values for binary inputs (R1 , R2) and the bias Memristance (R0) values are shown in the table, which have been found using MR-II and scaled to resistance values. The Memristance can be changed to program the circuit to perform different operations. The op-amps have a virtual ground reference of 2.5V and the input logic level is 0.1V above and below this level. The results for an AND and OR gates are shown below.

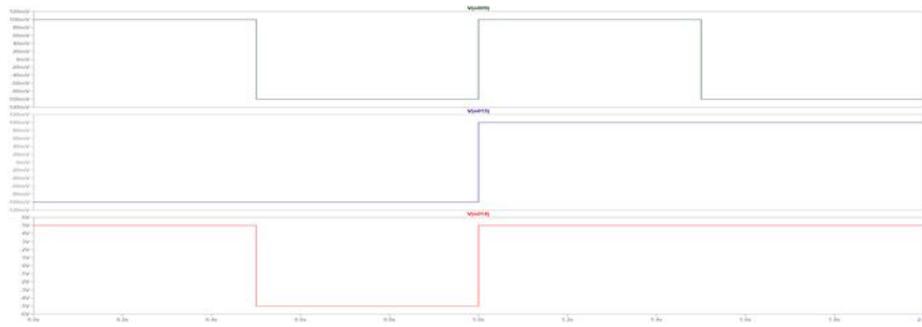

Fig.21a. OR Gate Output

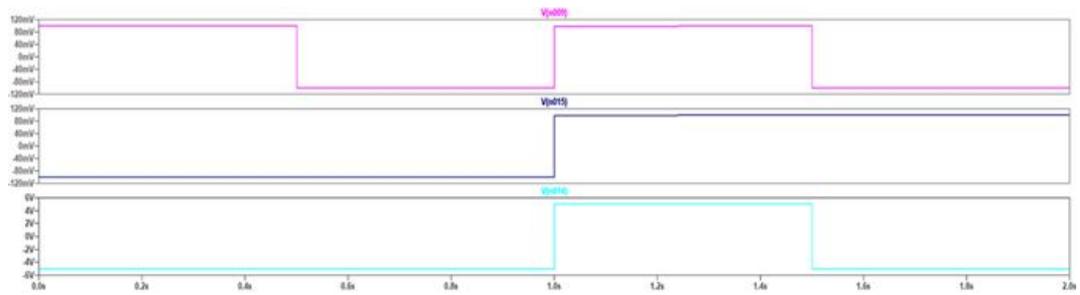

Fig.21b. AND Gate Output

# 3. Artificial Neural Network Building Blocks

### 3.1. Weight Block:

The weight block consists of a memristor bridge circuit connected to a differential pair. The bridge can be decomposed into a parallel combination of two series memristor circuits which contain two memristors in an anti-symmetric configuration. The series anti-symmetric configuration [33] has the property of constant resistance and the memristor bridge can easily be programmed through the input pin.

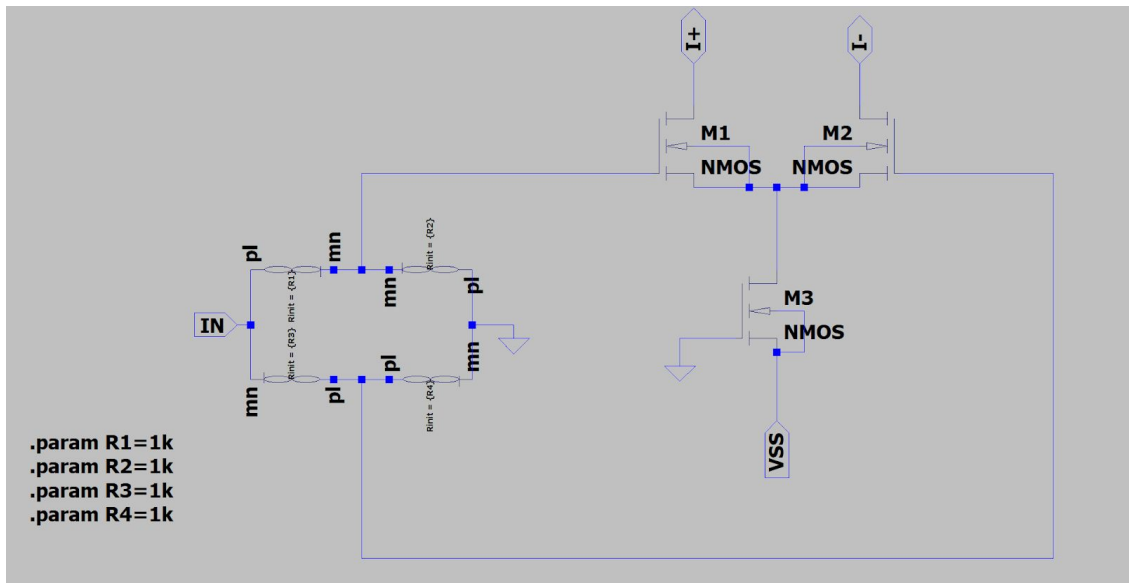

Fig 22. Weight Block Circuit Implementation

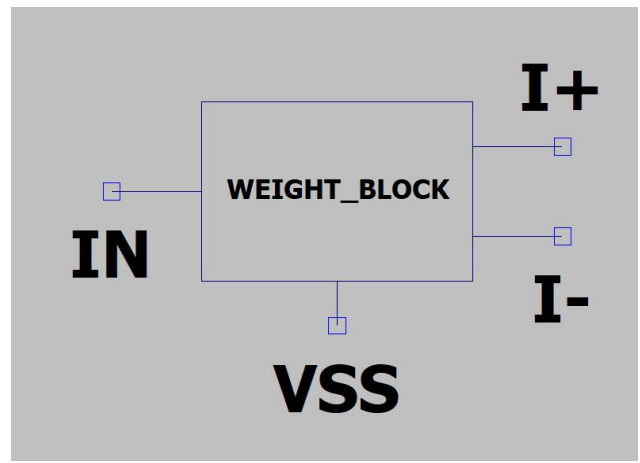

Fig 23. Weight Block Interface

The two output terminals I+ and I- feed to a summing block where the currents are subtracted. The weight block has a programmable weight ranging from -0.98 to 0.98 and can be easily programmed to have a zero weight by balancing the bridge.

### 3.2. Summing Block:

The summing block is responsible for summing the I+ and I- currents from the weight blocks and then subtracting their total sum. The summing block has been implemented using an OTA style circuit as shown below:

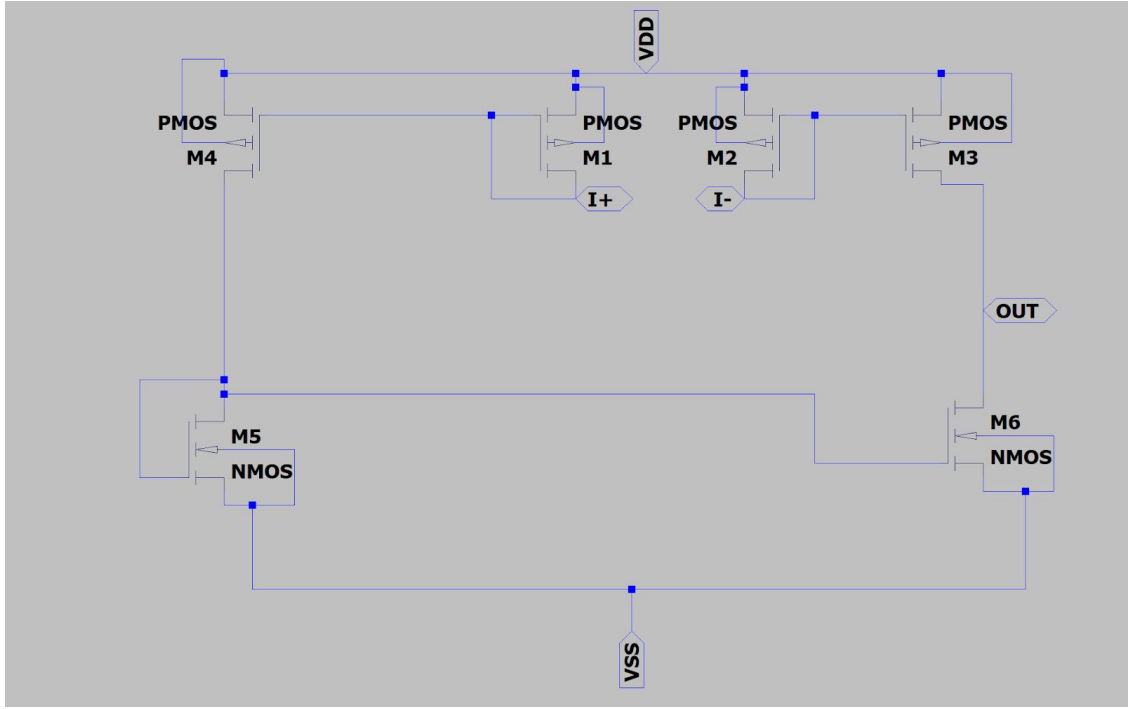

Fig 24. Summing Circuit Implementation

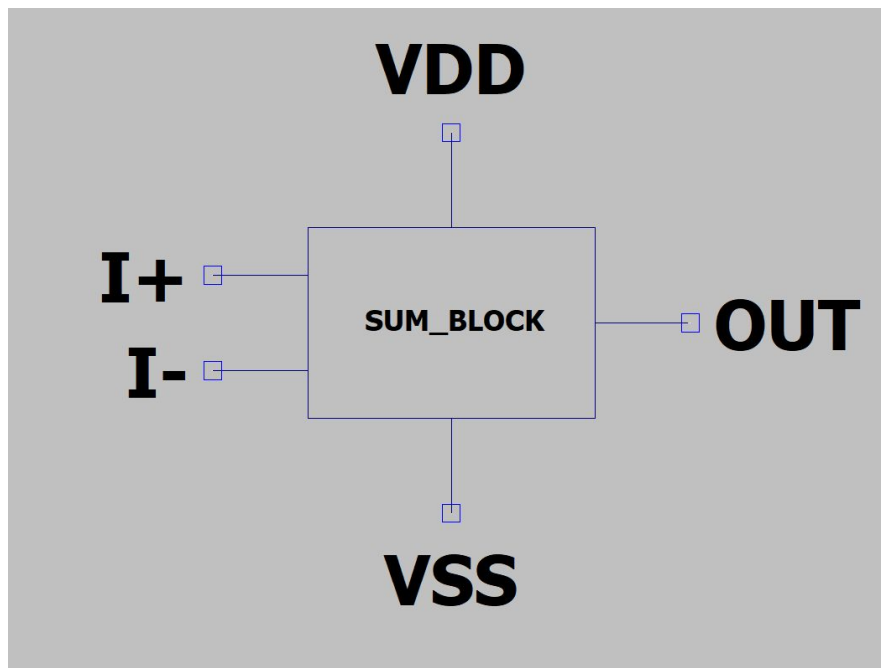

Fig 25. Summing Block Interface

### 3.3. Activation Function
### 3.3.1 Introduction

AFs decides whether the neuron fires or not for the computed weighing function and the equivalent biases. The parameter of the output data are produced in a manner such that, by incorporating some gradient algorithm such as the gradient descent algorithm. The AFs can be of different forms zero centered, linear or nonlinear. Depending on the application they are to be used, they are selected and used to control the outputs of out neural networks. These are used across different domains from speech recognition, object recognition and classification, scene understanding and description, cancer detection systems, fingerprint detection, weather forecast, self-driving cars, and multiple different domains. By categorizing the application of the activation function impact in various domains, recurrent analysis problem is reduced.

The output layer is preceded by multiple hidden layers which are processed by multiple linear weighted function to generate the final output. The input vectors a transformation is given by

$$f(a) = w^T a + b \qquad\qquad -(14)$$

where a = input, w = weights, and b = biases.

The neural networks produced from the Eq.(14) are further processed by not a single weighted input but by the matrix extrapolation of the weight vector with the input. Output of these models are given by from Eq. (15)

$$y = (w1.\, a1\, +\, w2\, .a2\, +\, ..\, +\, wn\, .an\, +\, b) \qquad\qquad -(15)$$

In order to prevent the problem of overfitting, special non-linear activation function are required to be applied to the input. These activation function are also called as transfer function and are utilized to modulate the input data. The final result after the application of the nonlinearity with the activation function is given by the equation:

$$y\, =\, \alpha(w1.x1\, +\, w2.x2\, +\, ..\, +\, wn.xn\, +\, b) \qquad\qquad -(16)$$

Where α is the activation function.

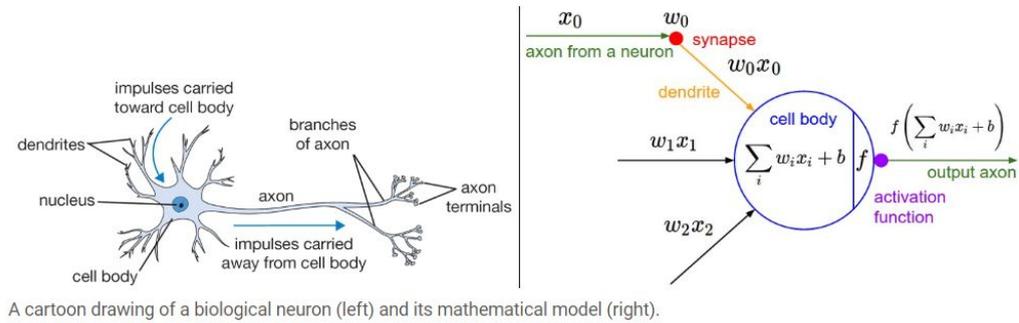

A cartoon drawing of a biological neuron (left) and its mathematical model (right).

Figure 26: Neural net Structure with an Activation Function {Source: CS231n Stanford 2017}

Some noteworthy points are:

1.      ***Location of the AF:*** For high output characteristics of the neural network, the AF should be placed at the point where to reduce the highly saturated output to a dignified scalable range, or to add specific non-linearity to the next stage to improve prediction.

2.      ***Zero Mean of data, using AF:*** The zero mean property of the activation functions like tanh and sigmoid enable the data to be re-centered about origin, preventing saturation of the output as it passed down to the next stage.

3.      ***Activation Function, effect on the gradient descent:*** Normally, the linear and the non-linear architectures work in a tandem in order to deliver to a particular task. Multiple problem come while designing the neural net such as the vanishing and the exploding gradient problem, of the derivative of the gradient term. In vanishing gradient, due to the repetitive multiplication of the gradient term, the net gradient vanishes to be zero. While on the other hand for gradient much greater than 1, it explodes towards infinity. The non-linearity due to the activation function limits the gradient terms in a defined range and prevents from these problems from occurring. Many early neural network activation function, somewhat biologically inspired were proposed by Elliott, 1993 as he studied the usage of the Activation functions in neural network.

### 3.2. Proposed CMOS based implementation of Activation Function

### 3.2.1. ReLU – Rectified Linear Unit

The characteristics in the designing part, which are to be aimed are:

$$f(x) = f(x) = \{0, \ x < 0 \ x, \ x \geq 0 \qquad -(17)$$

Research on the existing CMOS based circuits which offer similar characteristics was the point over which the exploration of design revolved. Current mirrors offer a fine solution to the above stated problem. The operational note on the current mirrors is discussed as below:

### 3.2.2. Operation Characteristics of Current Mirrors

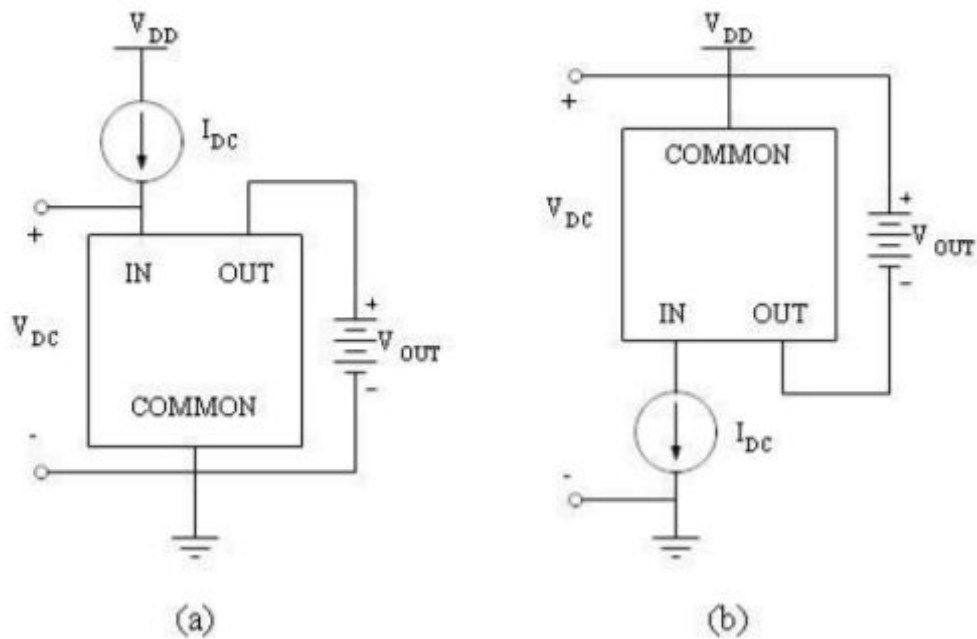

Figure 27: Current Mirror operation (a) NMOS based (b) PMOS based (Source: Lecture 3, EECS3611 Analog Integrated Circuit Design Course Material, York University)

The above figure basically depicts two models of the current mirror common in literature. One is the NMOS i.e. N-type Metal Oxide Semiconductor based current mirror and the other is the PMOS i.e. N-type Metal Oxide Semiconductor based current mirror. Both of these show the operational characteristics as:

$$I_{OUT} = I_{DC} \qquad\qquad -(18)$$

Multiple Implementation of the current mirrors are common in literature, some are summarized as:
1.      Simple Current Mirror (offers low output resistance)
2.      Cascode Current Mirror
3.      Low Voltage model of Cascode Current Mirror (Reduce minimum output voltage)
4.      Wilson Current Mirror (Feedback based operation)

### 3.2.3. Proposed Circuit for the ReLU Block

Cascode current mirror was chosen for the ReLU implementation because it offers a high output resistance. The circuit diagram of the cascode current mirror is shown in the Fig. 28.

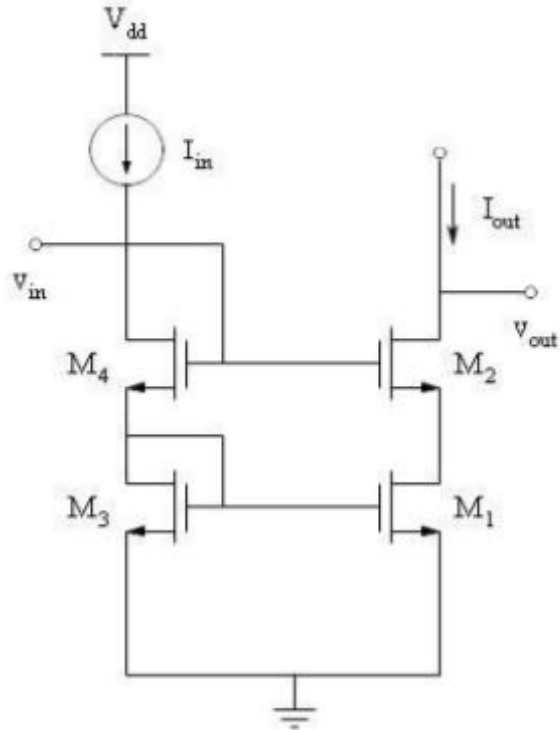

Figure 28: NMOS based Cascode Current Mirror (Source: Source: Lecture 3, EECS3611 Analog Integrated Circuit Design Course Material, York University)

The equation of the output current in the current mirror in the dc mode of operation is extremely simple. As, $V_{GS}$ is pushed equal on the both ends, leads to the same output current, considering the saturation mode of operation. The equation of whose current is given by:

$$i_D = k \, (V_{GS} - V_t)^2 \qquad\qquad -(19)$$

*where $V_{DS} > V_{OV}$ (i.e. $V_{GS} - V_t$) and $V t$ is the threshold voltage.*

Moreover, as $V_G$ and $V_D$ are joined the above condition is automatically satisfied, pushing them naturally in the saturation region.

If we consider the Fig. 28(a) part carefully, we can observe that the considering a singular orientation of the current direction, that the current direction is reversed over the next stage. Thus, giving the graph of a negative ReLU. In order, to solve this problem we add a NMOS current buffer after it, pushing the current in the original configuration. The limiting characteristics of the PMOS and NMOS to the current in source to drain and drain and source respectively are the underlying principles of the proposed research.

### 3.2.3.1. Simulation and Results

The circuit of a RELU based on cascode current mirror is shown in Figure 29.

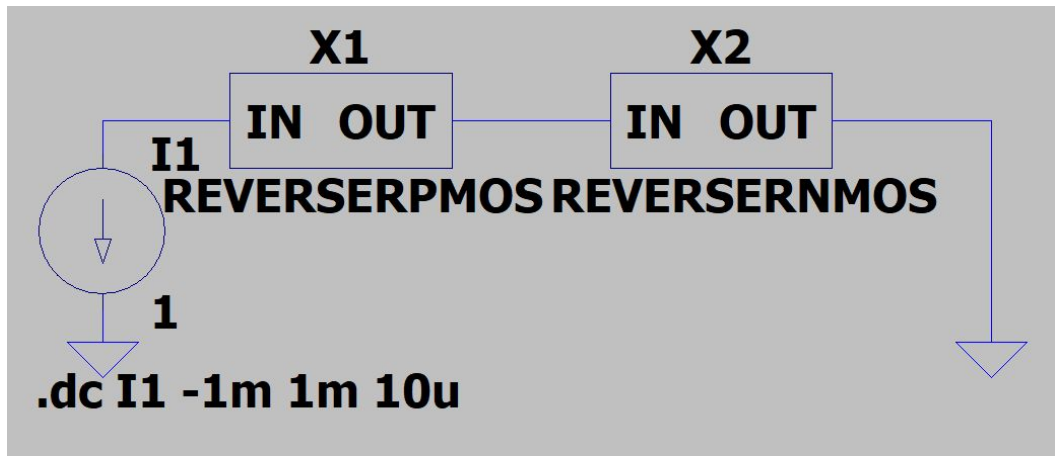

Figure 29: ReLU Implementation using Current Mirrors

Table 6: Simulation values for Current Mirror based ReLU

| Attribute | Value |
|---|---|
| Start (mA) | -1 |
| End (mA) | 1 |
| Increment (uA) | 10 |

The simulation results for the Cascode current mirror based configuration are shown in the following Fig, 30.

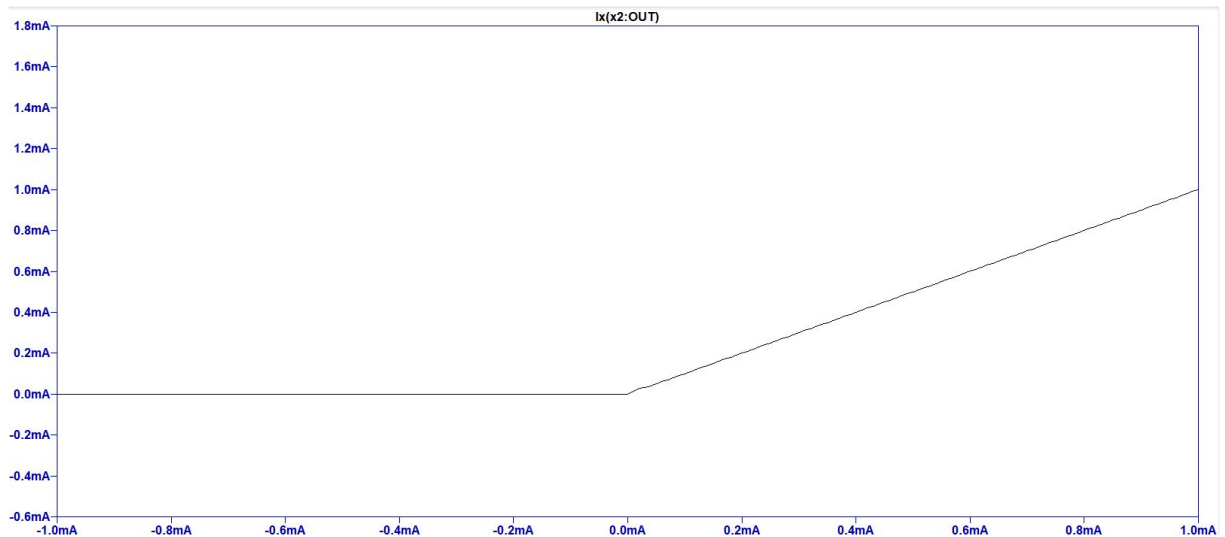

Figure 30: Simulation Characteristics of the proposed ReLU Model

### 3.2.4. Tangent Hyperbolic

Tangent Hyperbolic function, commonly known as tanh(x) is commonly, used for the implementation of the activation function, because of the zero mean property offered. This circuit offers an upper hand in the implementation in the hardware domain as the problem of the saturation of the outputs, causing an erroneous response is highly prevalent in the case of the hardware neural network accelerometers. The equation of the tangent hyperbolic function is given by Eq.(20).

$$tanh\,(x) = sinhx/coshx \;\; = \;\; \frac{e^x - e^{-x}}{e^x + e^{-x}}$$ -(20)

[39] proposed a circuit, based on weak inversion mode of operation of CMOS, by applying the translinear principle. However, there exists limited circuits that emulate this behavior in the saturation mode. The proposed research solves this problem.

### 3.2.4.1. Pade's Rational Approximation

The Pade's rational approximation is based on the factor of the continued fraction for the representation of the irrational functions, given in Eq.(21).

$$tanhx = \frac{x}{1 + \frac{x^2}{3 + \frac{x^2}{5 + \dots}}}$$ -(21)

Considering only the first denominator term of this approximation and approximating the terms below, $\frac{x^2}{5 + \frac{x^2}{7 + \frac{x^2}{\dots}}} \rightarrow 0$ we have the following approximation given in Eq.(22).

$$tanhx = \frac{3x}{x^2 + 3}$$ -(22)

### 3.2.4.2. Implementation of the Tangent Hyperbolic Approximation based on the Pade's Rational Approximation

The implementation revolves around constructing CMOS based logical block's for implementing the various mathematical functions like squaring, division and multiplication. The detailed implementation of each logical block are explained in detail in the following subsections.

### 3.2.4.3. Implementation of the Squaring Circuit

The squaring circuit aids in computing the denominator term i.e. $x^2$. The circuit utilized is adopted from [40]. As the ideal squaring of the $x$ when it is in the order of uA or mA leads to equivalent results in the order of pA or uA, which are unsustainable for practical application, a balancing division factor was added that can aid in the pushing the operation point to uA or mA. The circuit proposed in [40], does exactly this. The operating equation is given in Eq. (23).

$$I_{OUT} = \frac{x^2}{4 \times I_{ref}}$$ -(23)

where, $I_{ref}$ is the current adjustment factor and x is the input .

The LTSpice circuit diagram for the same is shown in the Fig. 31.

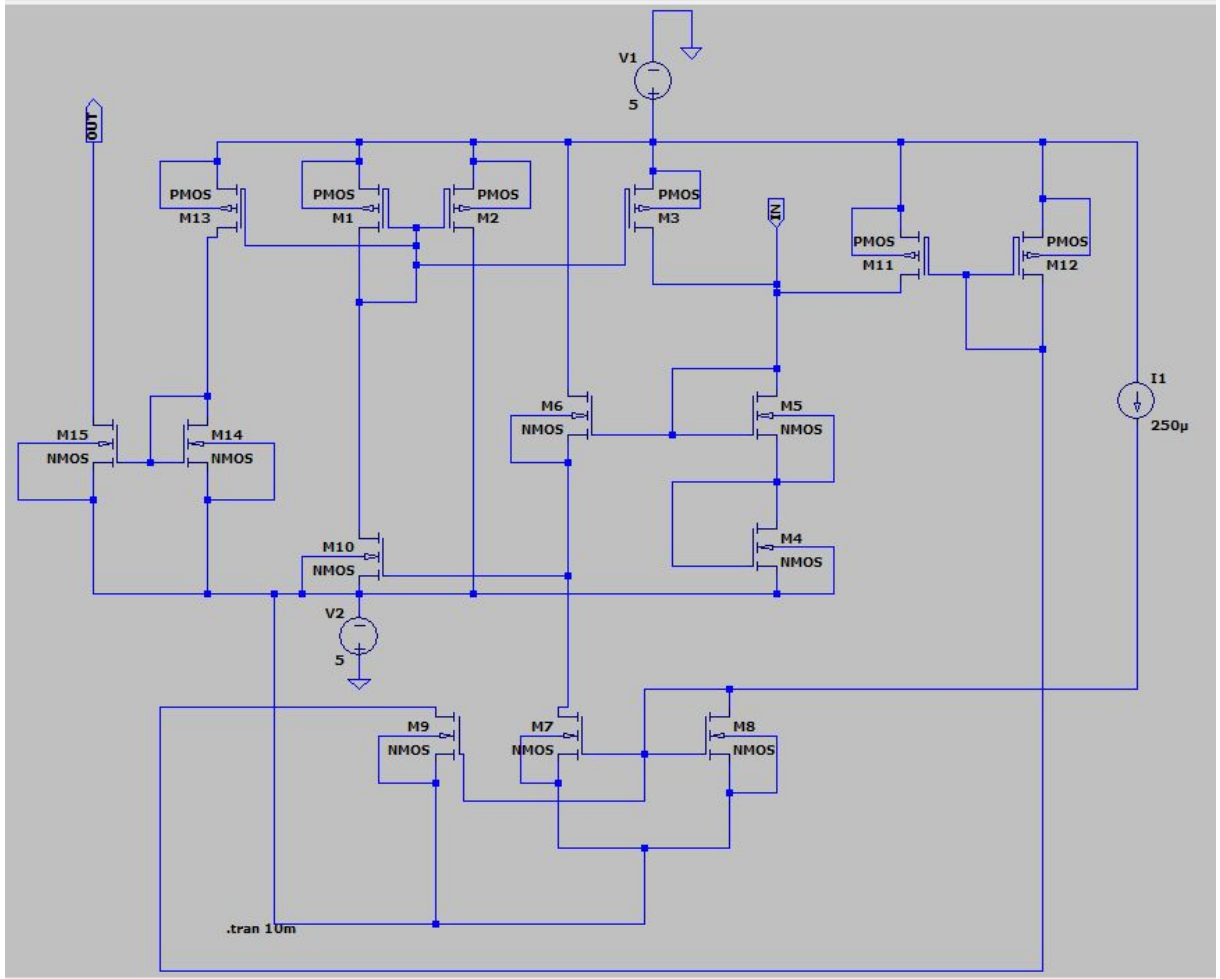

Figure 31 : Squarer Circuit Block

The value of $I_{ref}$ chosen is 250uA. This readjusts the current amplification factor to 1000 for the squarer.

$$I_{OUT} = 1000 \times x^2 \qquad\qquad -(24)$$

*where, x is the input current.*

However, the operating range of the multiplier is non-linear i.e. for certain value of input currents, it does not work. On applying DC Sweep on the multiplier for the values shown in the Table 7, the squaring operation is applicable only after the point where the input crosses the value of 360uA.

Table 7: Simulation settings for Squarer DC Sweep Analysis

| Attribute | Value |
|---|---|
| Start (mA) | 0.01 |
| End (mA) | 1 |
| Increment (uA) | 10 |

The characteristic curve for the squarer is shown in the Figure 32.

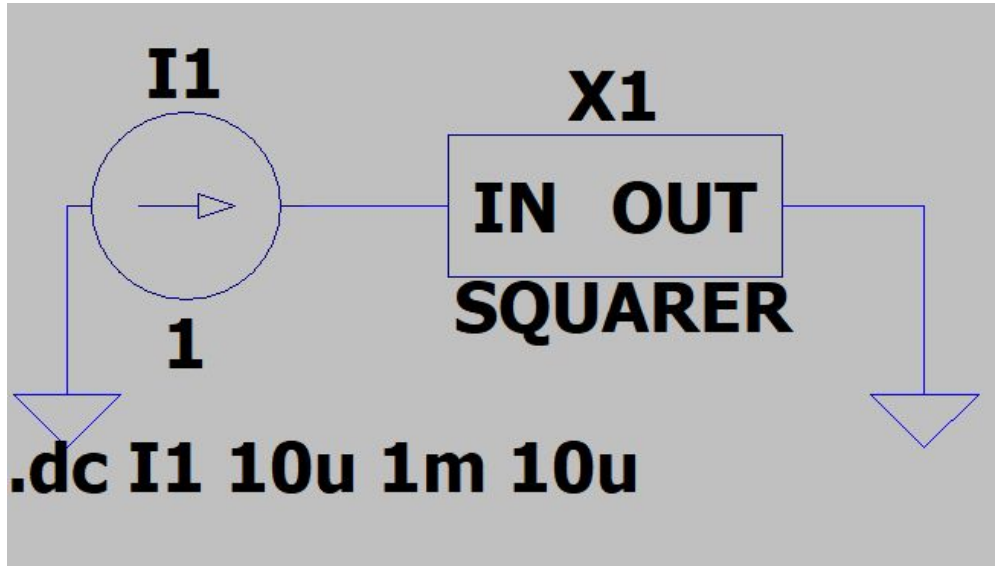

(a)

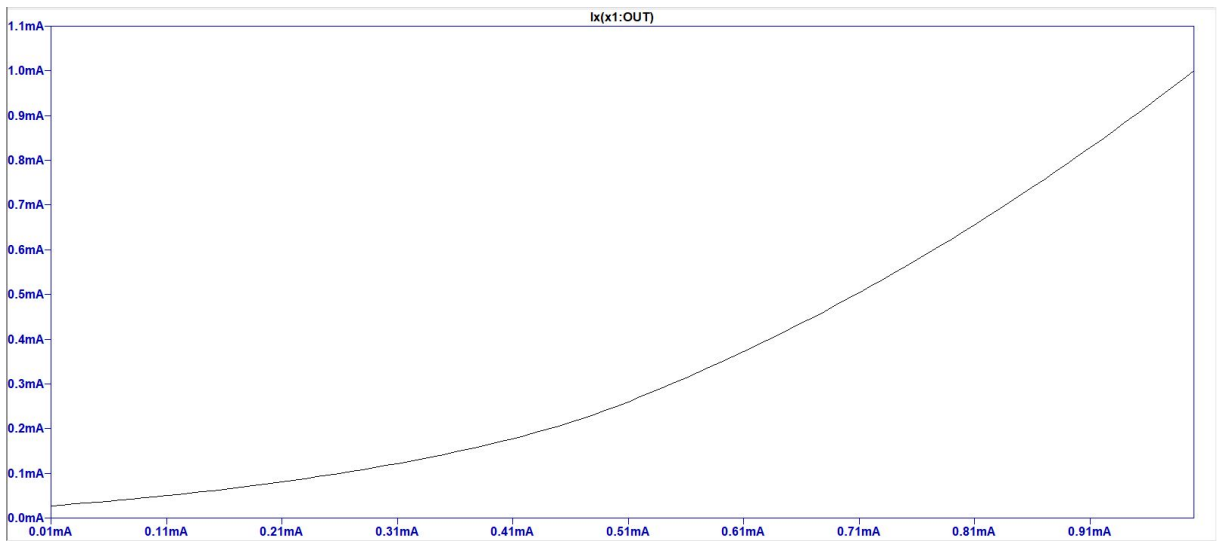

(b)

Figure 32: CMOS Squarer Characteristic Curve

### 3.2.4.4. Implementation of the Divider

The implementation of the current divider is based on the circuit developed in Section 1.2.2.3. However, in the case the $x$ considered in the Eq.(23) is considered to be a constant. While the $I_{ref}$ is the input $x$. The Eq governs the operation of the working of the divider circuit.

$$I_{OUT} = \frac{I_{REF}^2}{4 \times x}$$ -(25)

where, $x$ is the input current and $I_{REF}$ is the reference current.

Fig 33 shows the circuit diagram of the current divider adopted from [40].

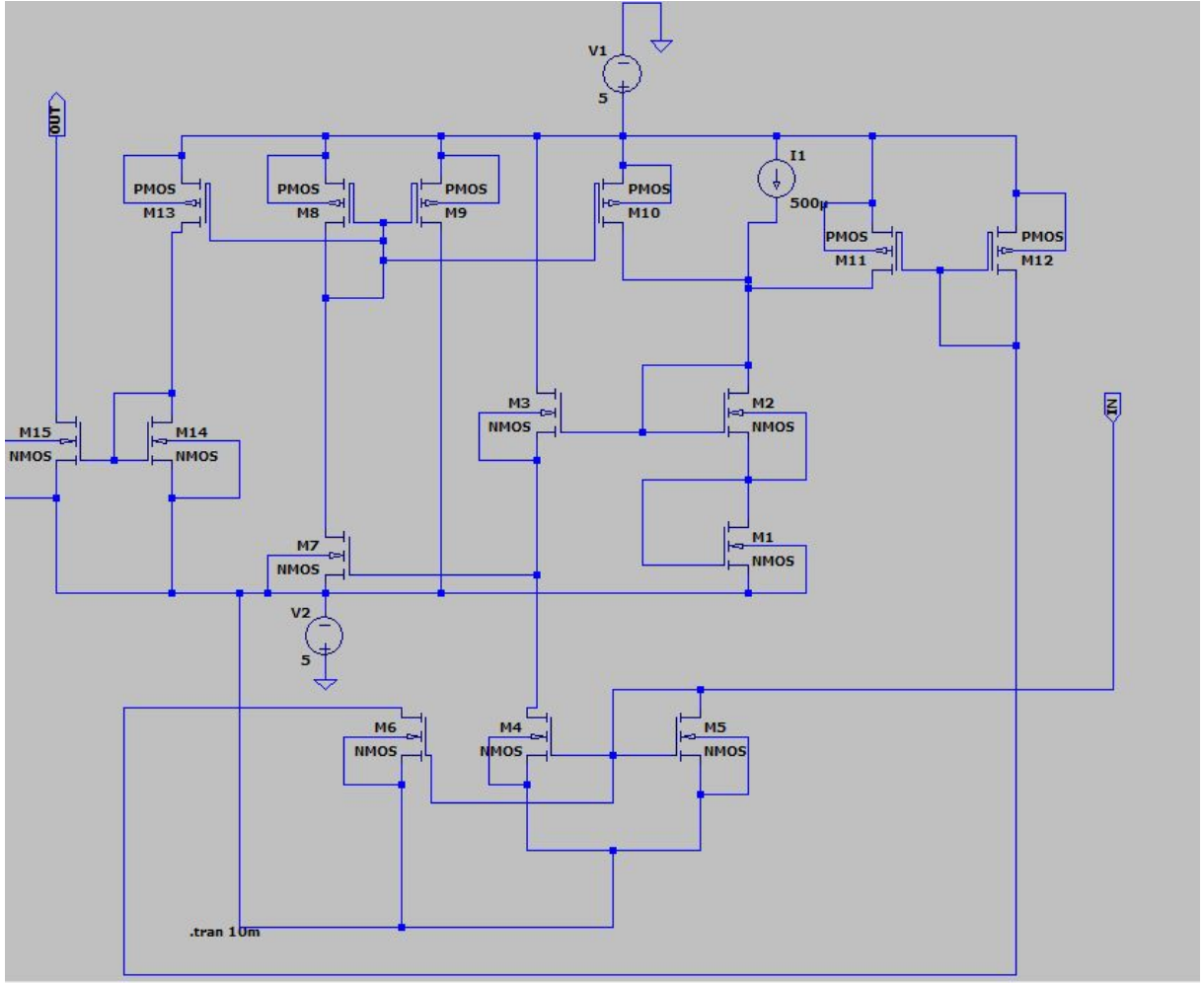

Figure 33 : Divider Circuit Block

The $I_{Ref}$ value adopted in the divider circuit is considering the viability of operation with other building blocks is 500uA. Thus, the resultant reduced equation is given in the Eq. (26).

$$I_{OUT} = \frac{1.5625 \times 10^{-8}}{x} A \qquad \text{-(26)}$$

where, x is the input current.

On the analysis of the divider circuit, we observe that for the values of $x > 300uA$, the circuit becomes non-operational. However, for values as less as 50uA, it works adequately. Considering these factors a suitable current adjustment methodology is required for re-adjusting the input current to the divider, which are discussed in 1.2.2.6.

The output characteristics of the Divider circuit is shown in the Fig. 34 for the DC Sweep settings given in Table 8.

Table 8: Simulation settings for Divider DC Sweep Analysis

| Attribute | Value |
|---|---|
| Start (mA) | 0.01 |
| End (mA) | 1 |
| Increment (uA) | 10 |

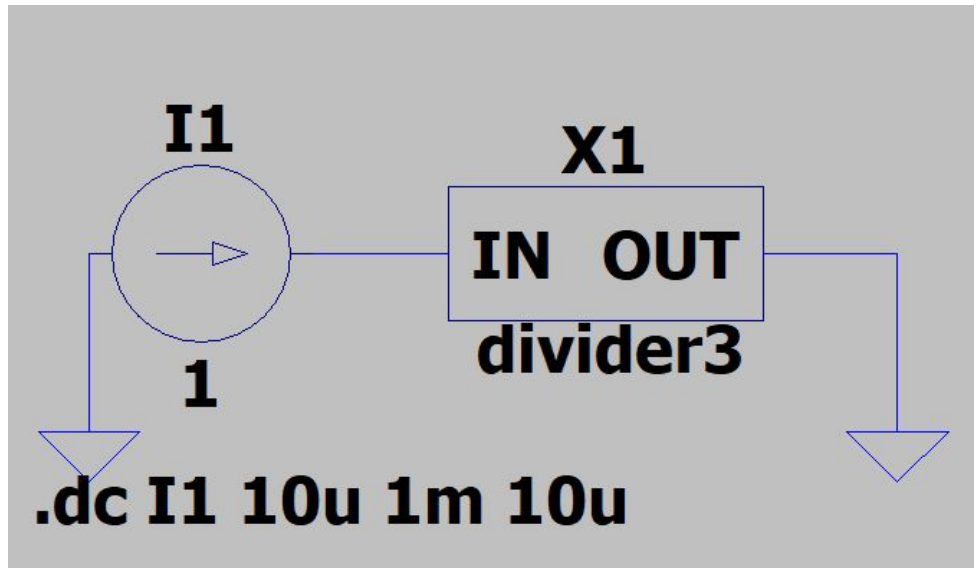

(a)

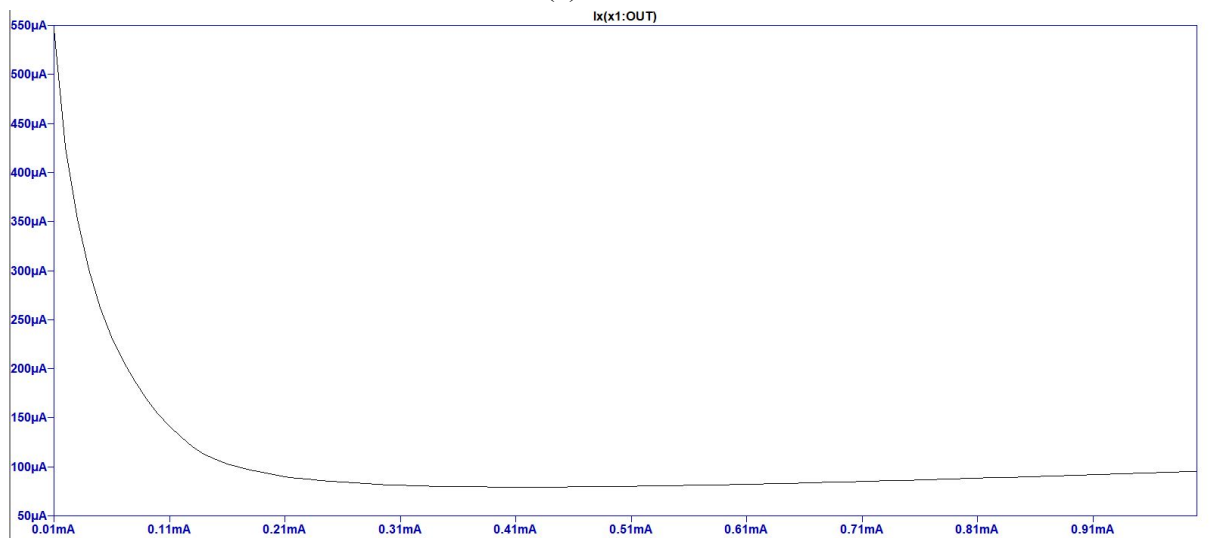

(b)

Figure 34: Output Characteristics of Current Divider

### 3.2.4.5. Implementation of the Multiplier

The multiplier implementation is based on the squarer circuit given in the Subsection 1.2.2.3. The implementation of the multiplier is based on the Eq. (27).

$$I_{OUT} = (I_1 + I_2)^2 - (I_1 - I_2)^2 = 4I_1I_2 \qquad \text{-(27)}$$

The algorithmic flow of the implementation of the same is shown in the Fig. 35.

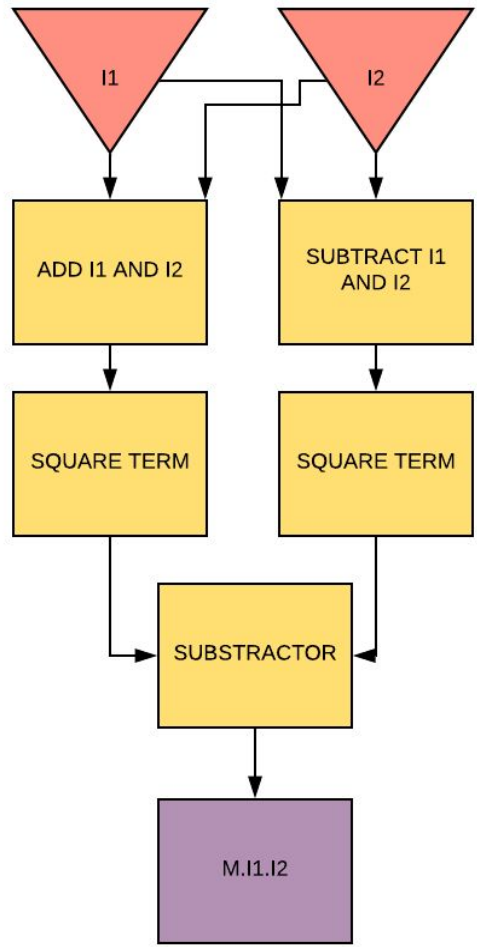

Figure 35: Algorithmic flow of Multiplier

Moreover, the following sub-circuits are required for the successful implementation of the above circuit using the squarer:

1.      Current subtractor circuit
2.      Interfacing Circuit
3.      Current Copier Circuit

*Current Subtractor Circuit:* The current subtractor unit used is the same that is utilized in the output stage of the memristor bridge programmable weight circuit.

*Interfacing Circuit:* The interfacing circuits are basically stages of NMOS and PMOS based current mirror for orienting the input current direction to the next operational stage. Here, majorly from the output of the subtractor to the input of the squarer, and following this again to the subtractor.

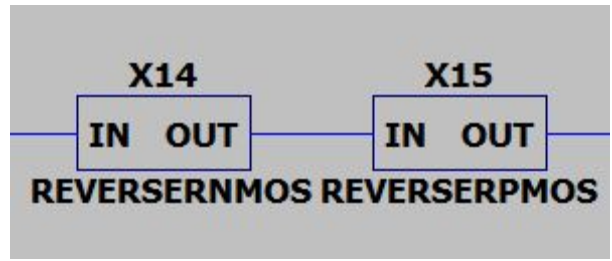

Figure 36: Interfacing blocks

**Current Copier Circuit**

The NMOS based current copier is utilised for feeding the input to multiple parallel stage. The implementation of the same is shown in the Fig 37.

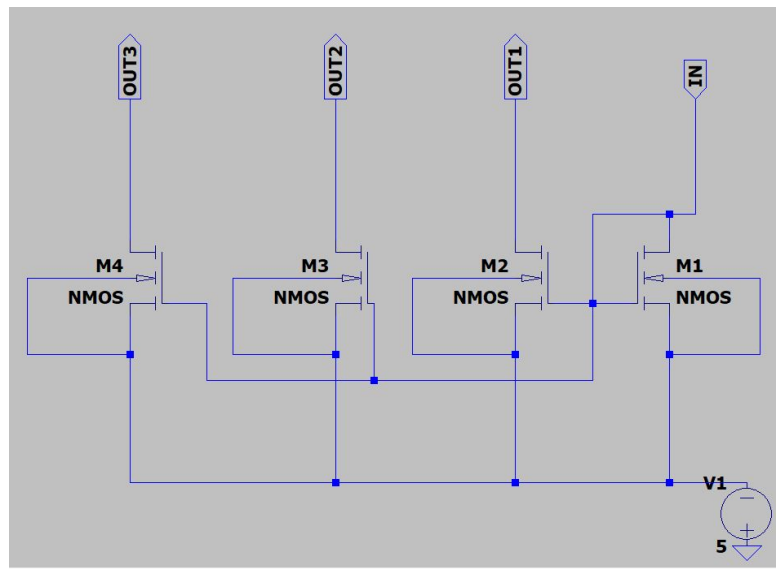

*Fig 37: Implementation of 3 way Current Mirror Circuit*

**Implementation of the multi-quadrant multiplication**

The multi-quadrant multiplication is implemented by entering the terns $(I_1 - I_2)$ and $(I_2 - I_1)$ to the squarer circuit. For the case of the negative input due to the biased nature of the NMOS, the other current output will tend to be zero. Thus, covering both the case scenarios. The block diagram implementation of the same is shown in the Fig. 38.

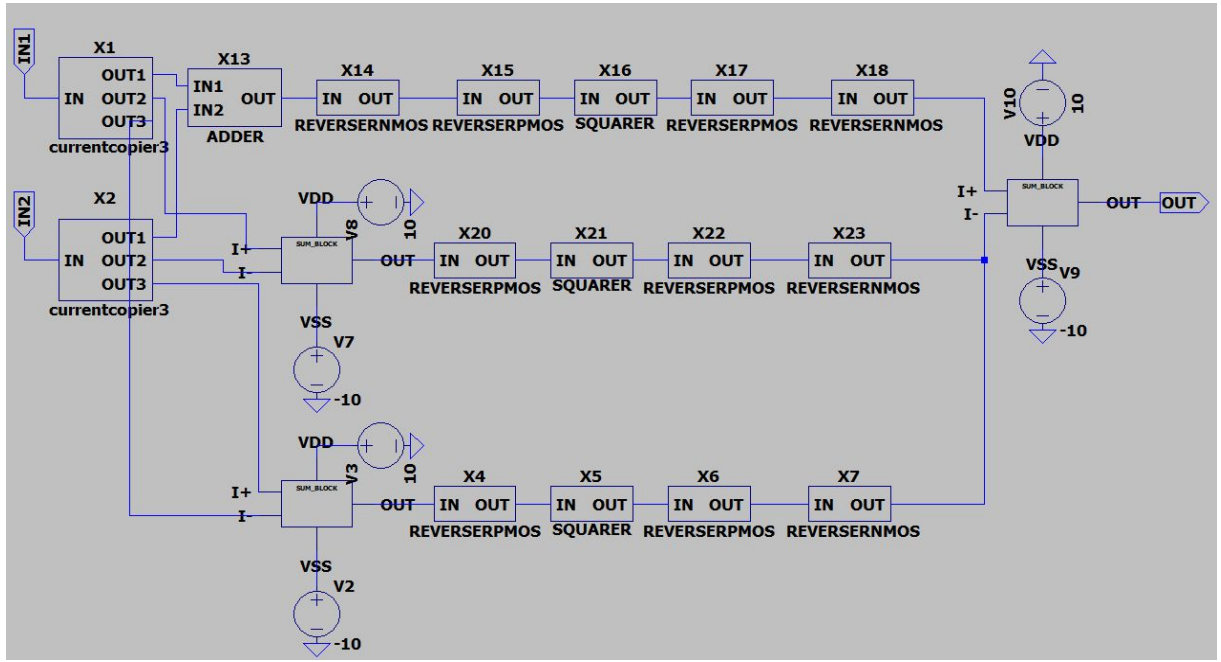

Fig 38: Implementation of the Current Multiplier Circuit

For the multiplier circuit, the equivalent multiplication equation is given in the Eq.(28).

$$I_{OUT} = 4000 I_1 I_2 \qquad \text{-(28)}$$

### 3.2.4.6. Interconnection and Calibration of logical blocks

***Interconnection of Elements:***

The problem of the limited current input range is solved by creating a 1/5 gain current mirror shown in the following Fig. 39.

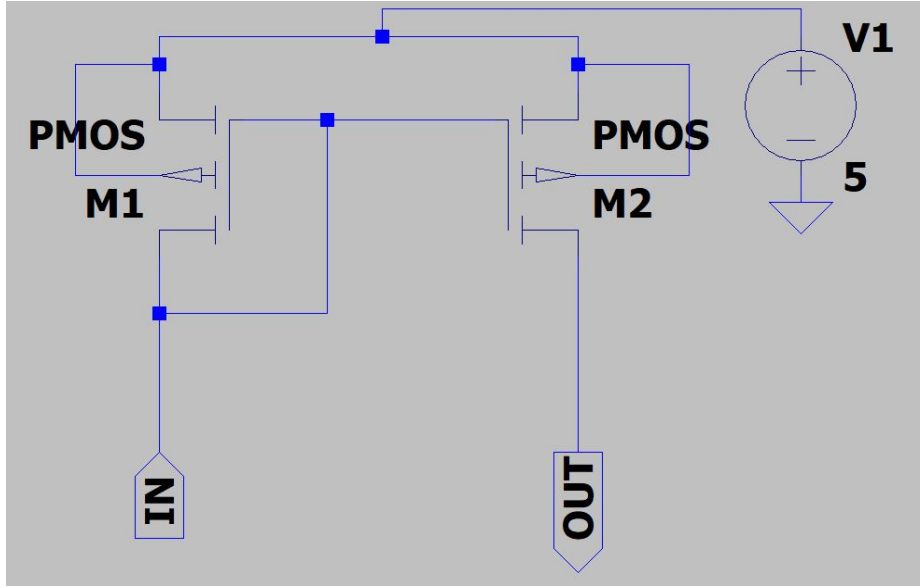

Fig 39: Unbalanced W/L ratio for limiting current

Table 9: W,L values for the current factor reduction

| Transistor | W,L |
|------------|-----|
| M1 | 180u , 0.18u |
| M2 | 32u , 0.18u |

The interconnection of the elements is done through the PMOS and NMOS buffer stage as shown in the Fig. 36.

***Calibration of Elements:***
Upon implementation of the different logical blocks the interconnection of the different stages is required. The equivalent expression will contain to multiplication factors of 4000 because of the elements 3, $x$ and the divided element. The division element will contain the factor of 1000 for the implementation of the term $x^2$ .

***Defining the standard units of x:***
A unit of 0.1mA is considered as the 1 equivalent scale on the x-axis for tanh(x).

Considering this factor and solving the following equation, we obtain:
$m = 120uA$ and $c = 0.3uA$ in the equation given in Eq.(29).

$$f(x) = \frac{mx}{x^2 + c}$$   -(29)

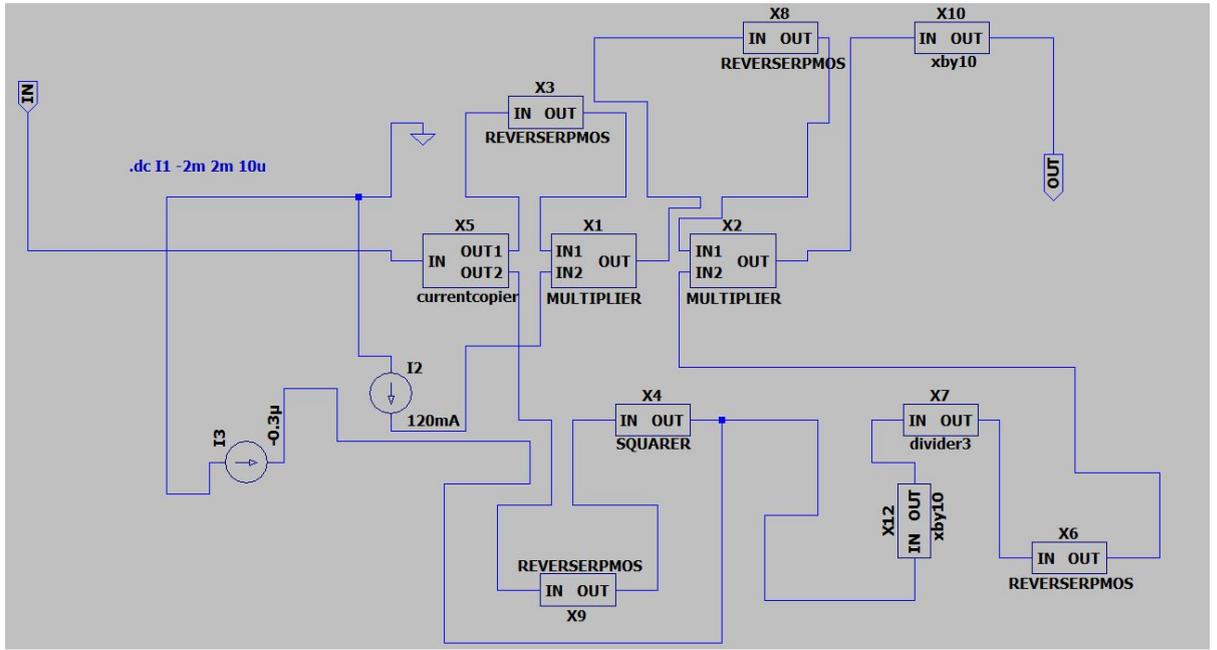

Fig 40: Implementation of the Tangent Hyperbolic

### 3.2.4.7. Simulation and output characteristics

The simulation of the circuit shown in Fig 41(a) was done. The simulation settings for the DC Sweep of the current source I1 is given in Table 10.

Table 10: Simulation settings for Tangent Hyperbolic

| Attribute | Value |
|-----------|-------|
| Start (mA) | 0.01 |
| End (mA) | 2 |
| Increment (uA) | 10 |

The output characteristics of the same are shown in the Fig. 41(b).

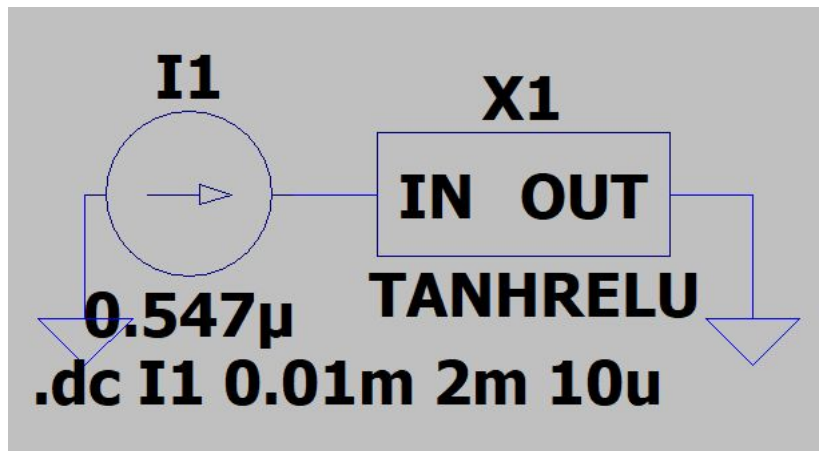

(a)      Circuit Diagram

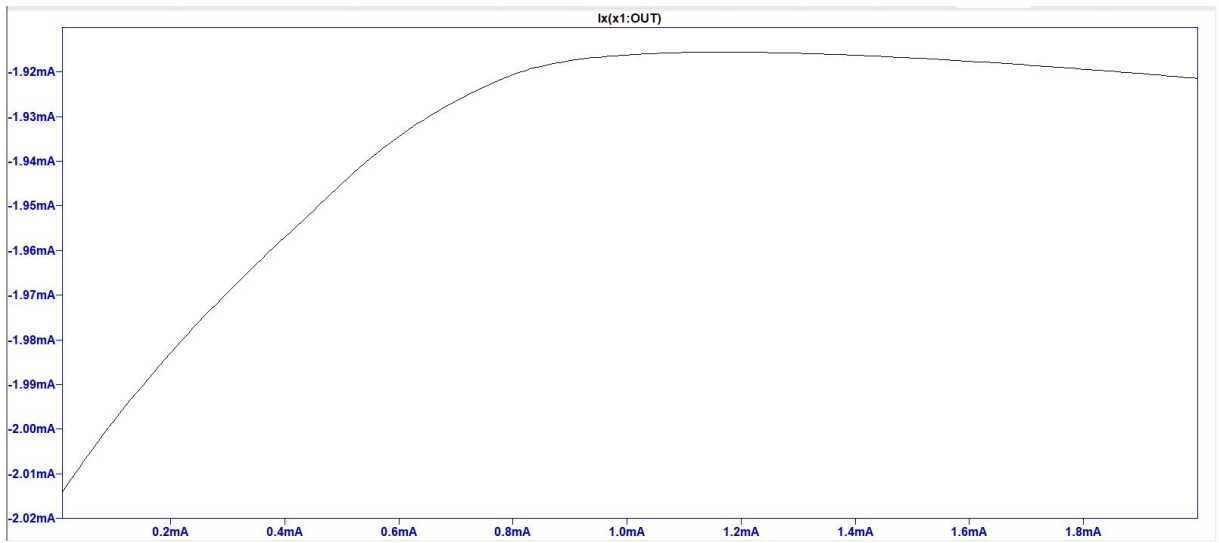

(b)     Simulation Results

Fig 41: Output characteristics of the Tangent Hyperbolic

### 3.5. Artificial Neural Network Implementation:

To test the feasibility of the blocks described, we have used them to construct a 2-layer neural network and compared its performance with the software implementation of the same network. This is a fully connected network with each node connected to the nodes in the layer before and after it. We have used the patternnet architecture from the neural network toolbox in MATLAB and modified the activation function before training the network. The neural network structure is shown below:

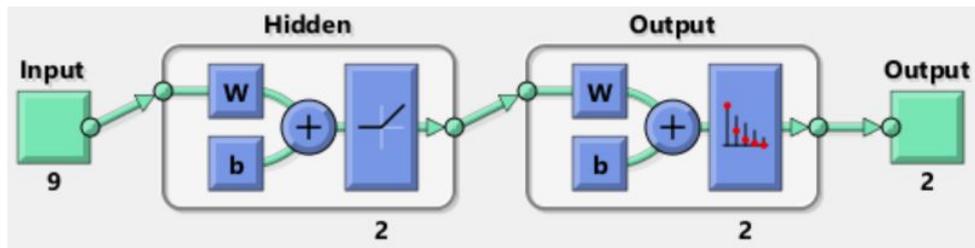

Fig 42: Patternnet Architecture Used {Source:Generated by MatLab}

The network consists of nine inputs which are fed to each of the nodes of the first hidden layer. The first hidden layer uses a rectified linear unit as the activation function. The output of the first hidden layer is fed into each of the nodes of the next layer. The output is then passed through a softmax layer which converts the inputs into two distinct probabilities which gives an estimate of the confidence of the neural network in classifying the input as each class. The network was trained using scaled conjugate gradient backpropagation. We have used 2 nodes in the first hidden layer to decrease the size of the network.

### 3.5.1. Dataset Used:

The neural network used the Breast Cancer Dataset available at the the UCI Machine Learning Repository [36]. The dataset contains information regarding the biopsies of tumors stored as 9 features:

1. Marginal Adhesion
2. Bare nuclei
3. Uniformity of cell shape
4. Clump thickness
5. Single epithelial cell size
6. Uniformity of cell size
7. Mitoses
8. Normal nucleoli
9. Bland chromatin

The data is normalised and fed into the neural network to classify the tumour as either benign or malignant. The data was split into 70% for training the neural network, 15% for validation and 15% for testing.

### 3.5.2. LTSPICE Implementation:

Using the blocks we have developed before, the circuit implementation of the neural network was developed. The weights of each layer from the software implementation were used to program the memristors in each of the weight blocks in the circuit. The complete circuit is shown below:

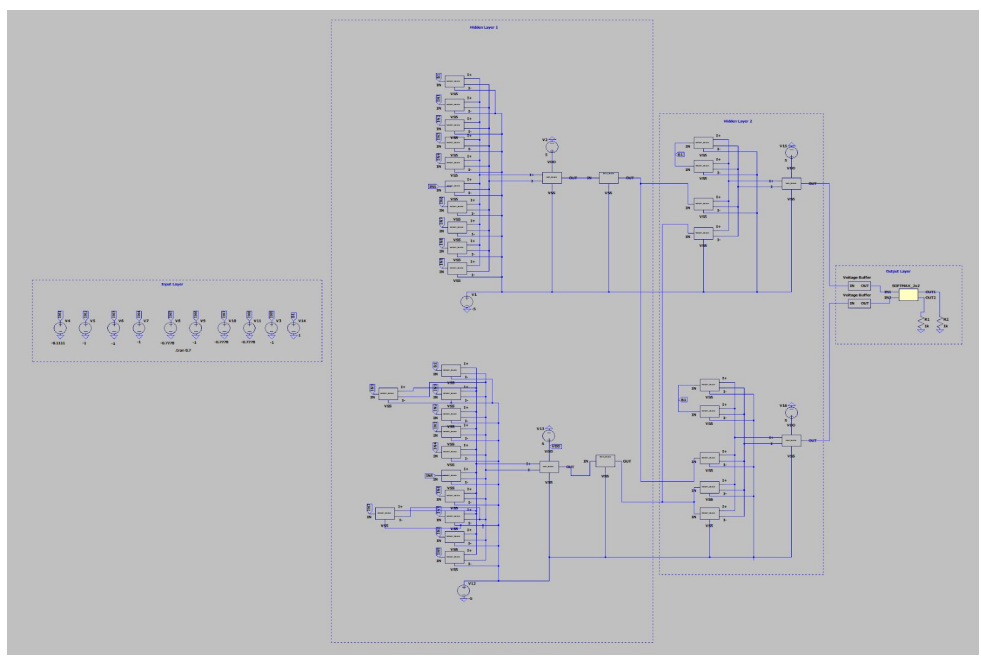

Fig. 43: Complete ANN Architecture SPICE implementation

### 3.5.2.1. Input Layer:

The input layer consists of a nine input signals which represent the nine parameters obtained after doing a biopsy on a tumour. The nine parameters have been normalised and zero-centred before being converted into a voltage signal.

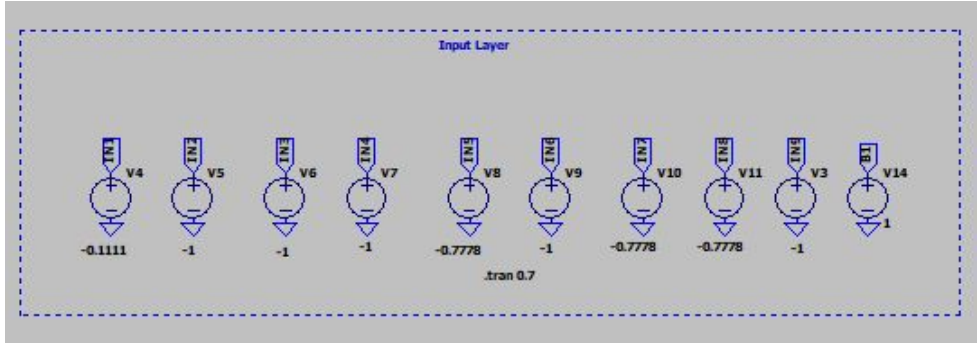

Fig 44: Input Layer to the ANN

### 3.5.2.2: Node Implementation:

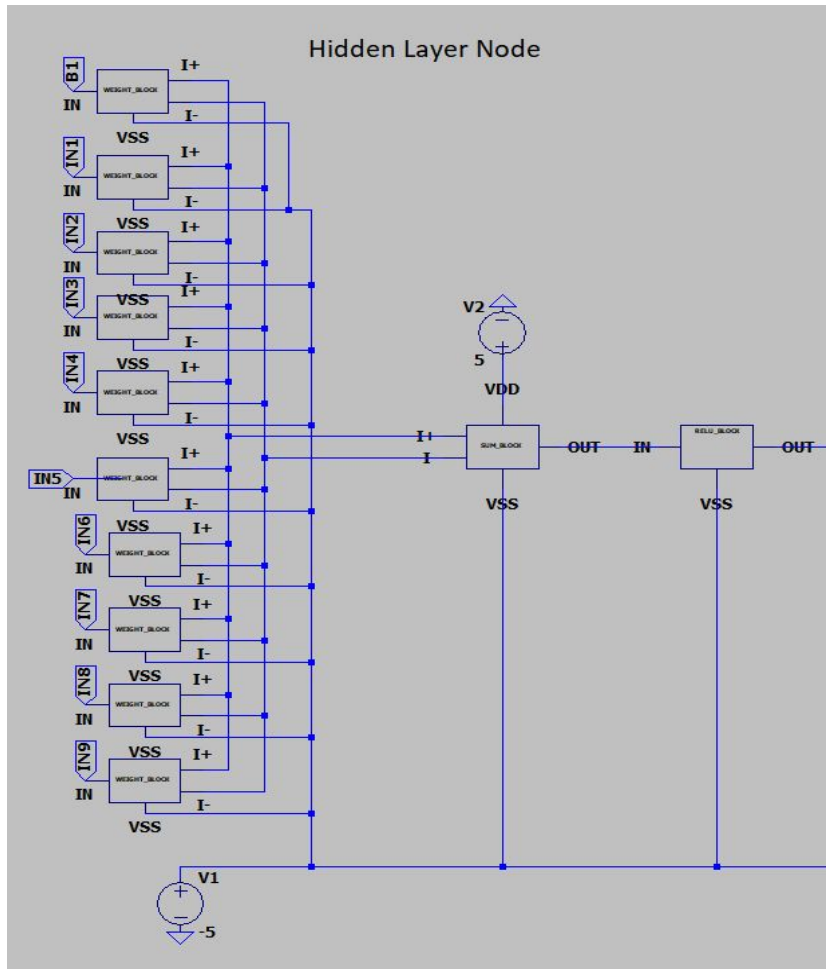

Fig 45. ANN Node Implementation

The neural network node is constructed using weight blocks, a sum block and an activation block. The weight block consists of a memristor bridge circuit which multiplies the input signal with the programmed weight in the memristor. The 9 input voltages require 9 weight blocks and the result of the multiplication of the signal and weights are accumulated using the sum block. The activation block present in the above node uses the ReLU (Rectified Linear Unit) activation function to introduce non-linearity between the layers.

### 3.5.2.3. Output Layer:

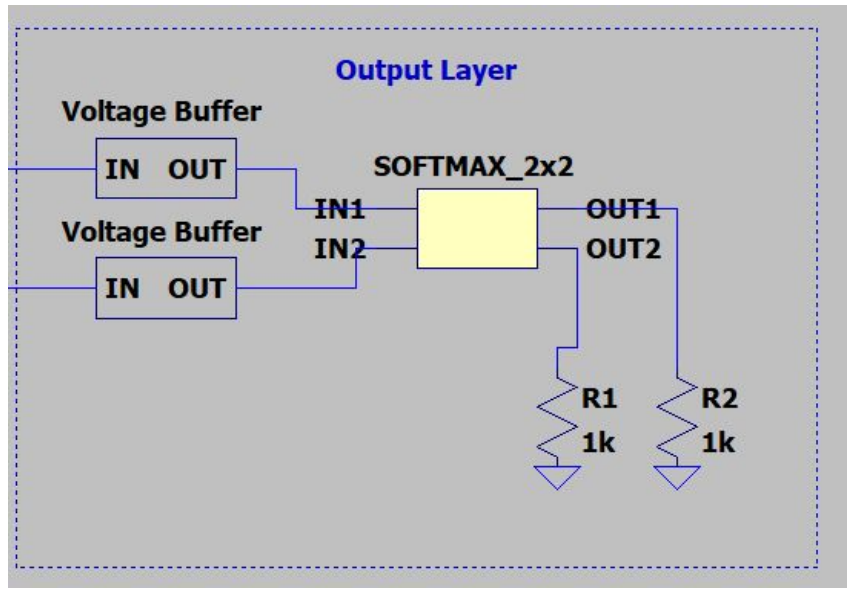

Fig 46. Output Layer

The output layer has a voltage buffer connected to a softmax block which has been implemented using a sub-circuit netlist using a dependent source and inbuilt SPICE mathematical functions. The softmax function basically performs exponential averaging.

$$e^{-z1}/\prod_{i=1}^{2} e^{-zi} \qquad\qquad -(30)$$

The output from the softmax block is the probability of the tumour being benign and the second output terminal gives the probability that the given tumour is malignant.

### 3.4.3. Performance Comparison:

The artificial neural network has an accuracy of 96.71% in classifying the data with the given architecture. The SPICE implementation performs along similar lines with a slightly lesser accuracy of 94.85% in classifying the data.

The network was tested with sample data given in the form of input at 1 millisecond intervals. The outputs at the softmax layer are shown in the figure below.

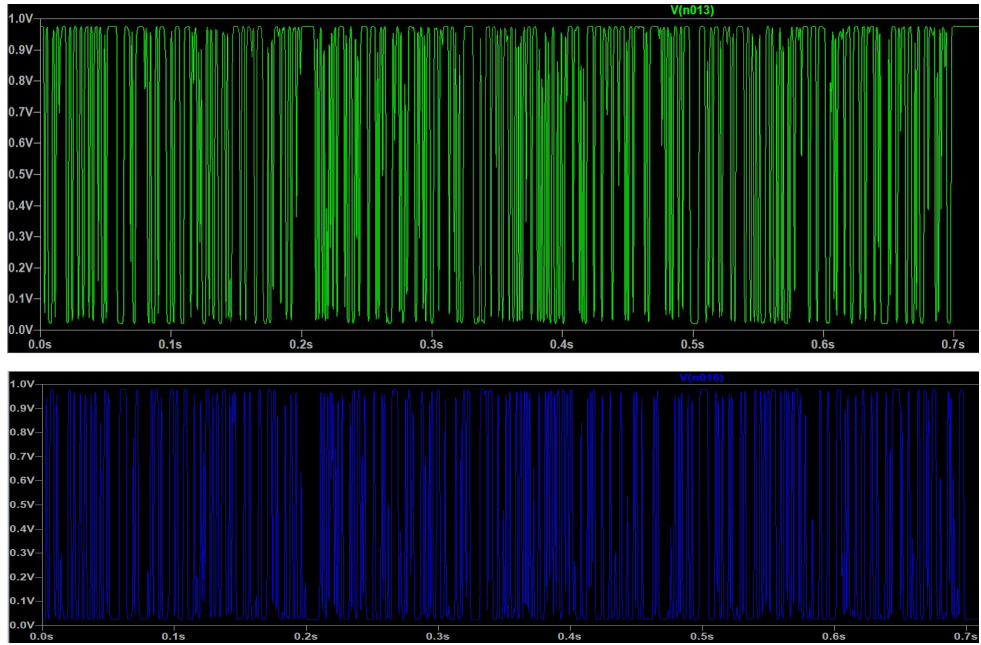

Fig 47. Output of Softmax Layer

# 4. Convolution Neural Network Blocks

## 4.1. CMOS-Memristor hybrid implementation of convolutional filter:

Image kernels, are convolution based image processing breakpoints that enable sharpening, blurring, edge detection in the input image. The application of these results are wide, ranging from obstacle detection in drones to real-time image enhancement in a drone on, say mars. Moreover, the proposed circuitry enables the weights in the kernel to be floating point, without any additional processing as in the digital architectures.

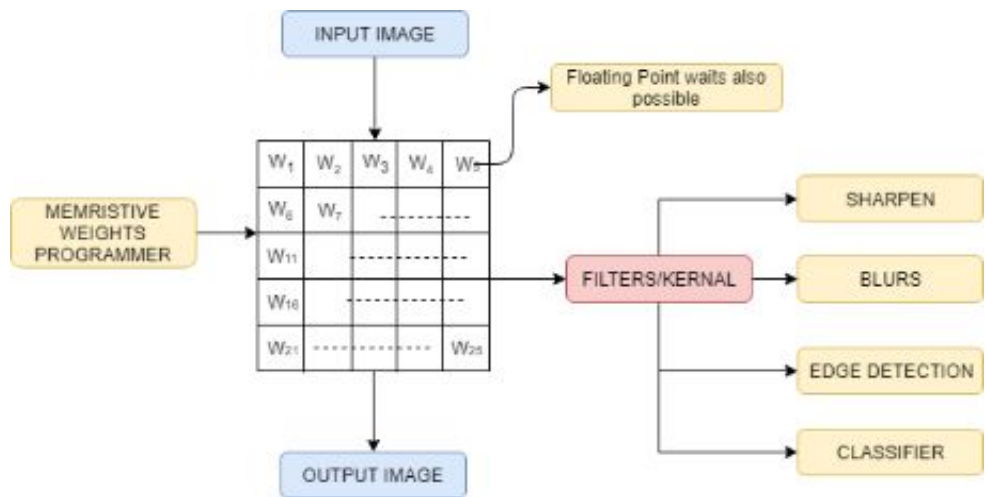

Fig 48 : Block diagram of memristor-based programmable image kernel CNN implementation

The proposed algorithmic flow for our application is shown in Figure 42**.** The major building blocks for developing this application includes:

1.      Method for handling programming for synaptic weights in the circuitry.

2.      Circuitry for parsing the required voltage levels over the multiple sections of voltage inputs to be directed to the voltage multiplier and addition.

3.      Handler for synaptic weight addition and integration of simpler CMOS logics for specific application.

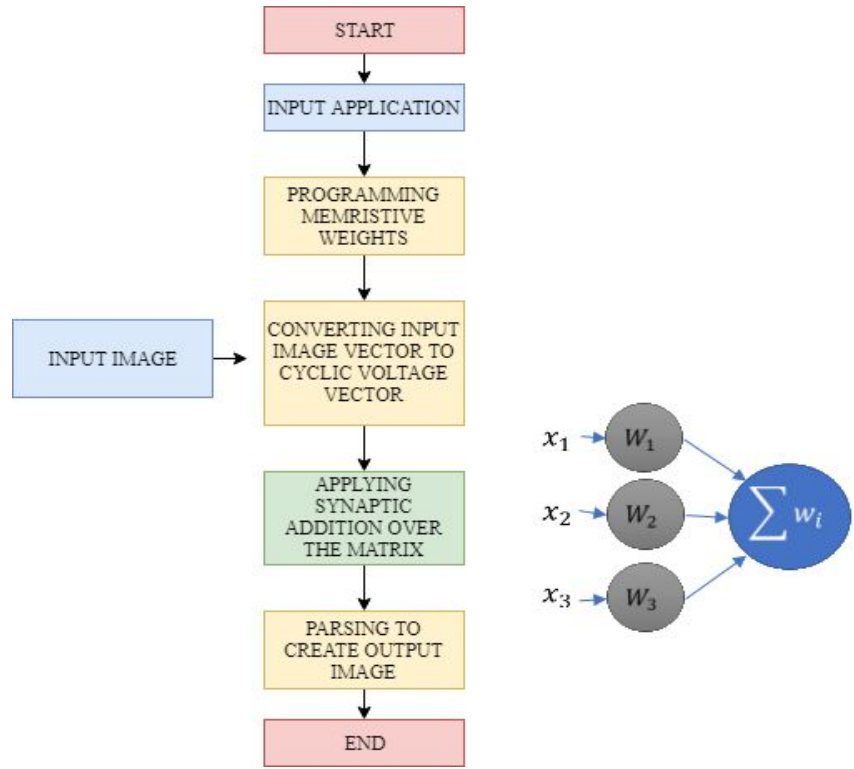

Fig. 49 : Algorithmic flow for image kernel programming

Using CMOS-Memristor hybrid circuits we implemented a 3X3 kernel sized convolutional filter, thus at one time it can process 9 pixel values. Output is taken as voltage at the load block.

$$V_{load} = ( \sum w_i * V_{ini} ) * R_{load}$$

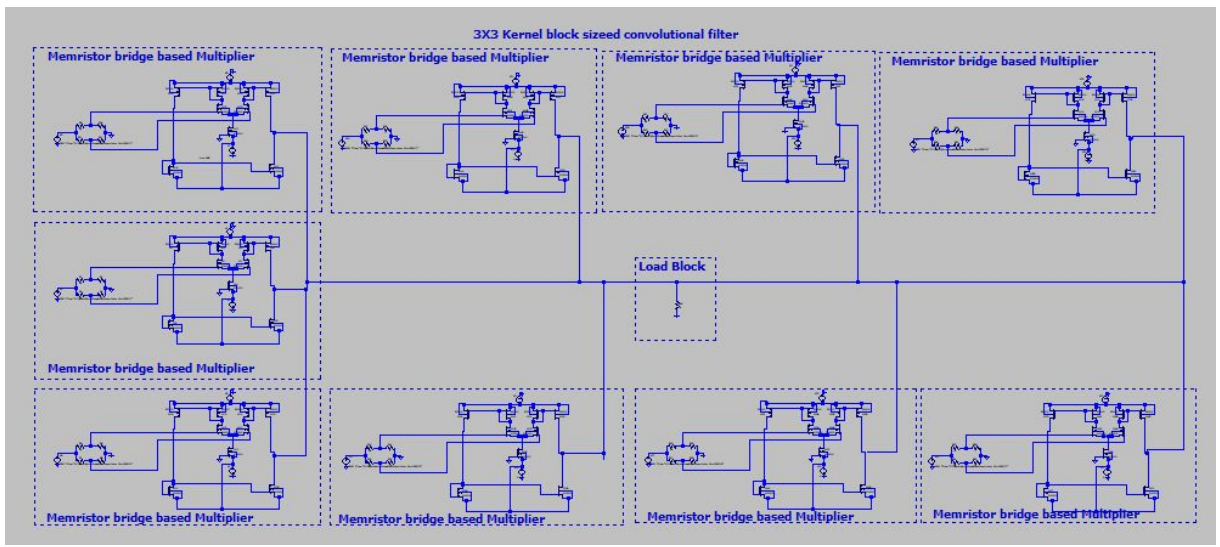

Fig 50 : 3X3 CMOS-Memristor hybrid kernel convolutional filter

The circuit comprises of 9 kernels each with programmable weights and the output of each kernel is connected together through a load resistor of 1K. Programmable weights are implemented through a Memristor bridge circuit along with a differential pair and active load. Ends of differential pairs are connected together with the active load so that current output from each kernel adds up.

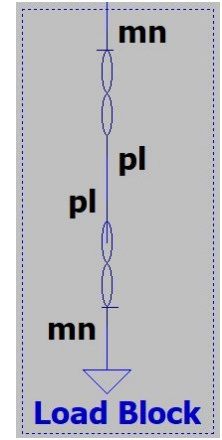

Fig 51: Load block

Voltage difference Vd is applied to the differential pair with an active load.

$$Vd = (\frac{M2}{M2+M1} - \frac{M4}{M4+M3}) * Vin \qquad -(31)$$

M1, M2, M3 and M4 are all memristors attached with a programmer circuit[38]. These memristors can be programmed to give both positive and negative weights between -1 and 1. Since, for a memristor practical values of Roff is 81K and Ron is 1k, total weight that can be obtained is limited to 0.98 to -0.98.

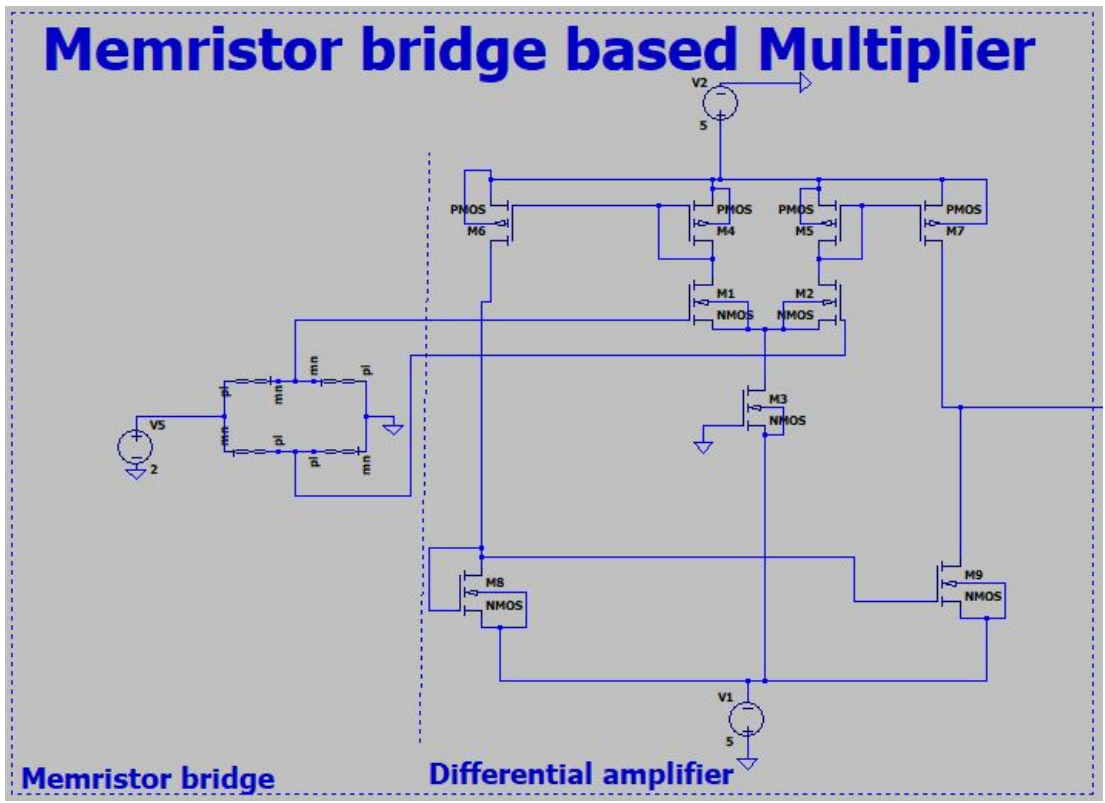

Fig 52: Memristor bridge based multiplier

To input an image, using MATLAB we converted the image to grayscale and then mapped the 8 bit pixel values (0 to 255) into a voltage signal of 0V to 1.5V. This voltage signal is then applied as a piecewise voltage source(in the form of .PWL file) for each multiplier. In a similar manner, output voltage between 0 to 1.5V is mapped to corresponding pixel values and then the output image is formed using MATLAB's *imread* function.

Using this method 2 filters were implemented:

**1. Average Blur filter**
In average blur filter, we took a kernel of weights [0.1, 0.1, 0.1; 0.1, 0.1, 0.1; 0.1, 0.1, 0.1]
Thus each memristor bridge is programmed to have these weights.

**Input image:**

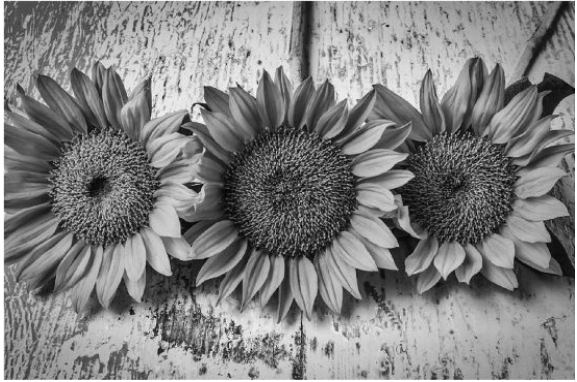

Fig 53a : Image input to CMOS-Memristor based blur filter

**Software output Image:**

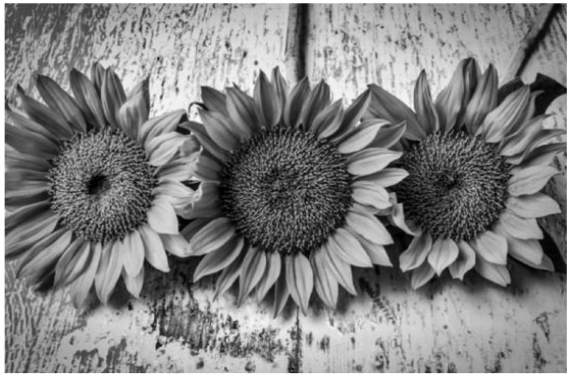

Fig 53b : Image output from software simulation of image kernels

**Hardware Output Image:**

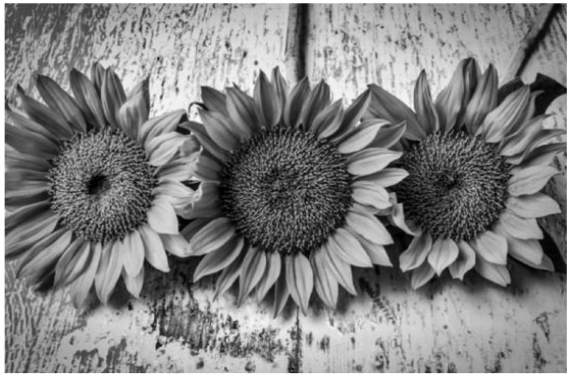

Fig 53c : Image output from CMOS-Memristor based blur filter

## 2. Edge detection filter

In edge detection filter, we took kernel weights of [-0.1 -0.1 -0.1 ; -0.1 0.8 -0.1 ; -0.1 -0.1 -0.1]

**Input image:**

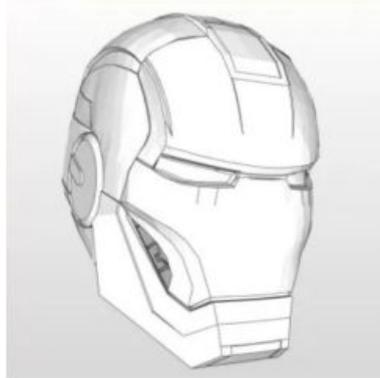

Fig 54a : Image input to CMOS-Memristor based edge detection filter

**Software output Image:**

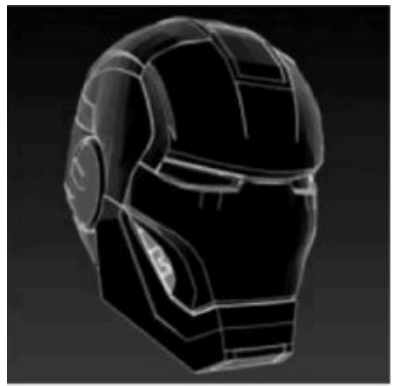

Fig 54b : Image output from software implementation of image kernels

**Hardware Output Image:**

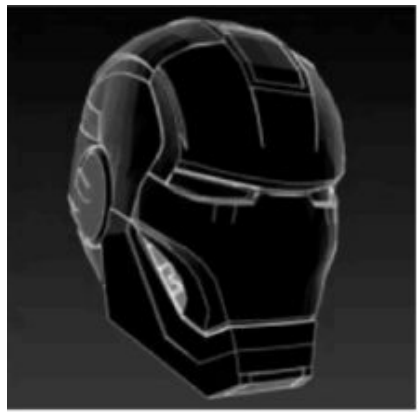

Fig 54c : Image output from CMOS-Memristor based edge detection filter

With these results, we were able to confirm the working of our programmable image kernel implementation.

## 4.2. POOLING LAYER

Convolutional layers outputs the feature sets of an input image but the problem with this feature set is that it is strongly linked with the position of the object within the image. We need out feature sets to be "Local translational invariant" i.e. independent of the position of the object in the image. One approach to do this is by downsampling the feature set and that can be done by pooling layer.

There can be 2 types of pooling layers: Max and average pooling.

Pooling layer is applied to the output of convolutional layer after the non-linearity(ex. ReLu). Size of the pooling layer is usually 2X2 applied with a stride of 2 units. Thus it will always reduce the size of the feature set to half.

There is another type of pooling called global pooling, which down samples a entire feature set to a single value.

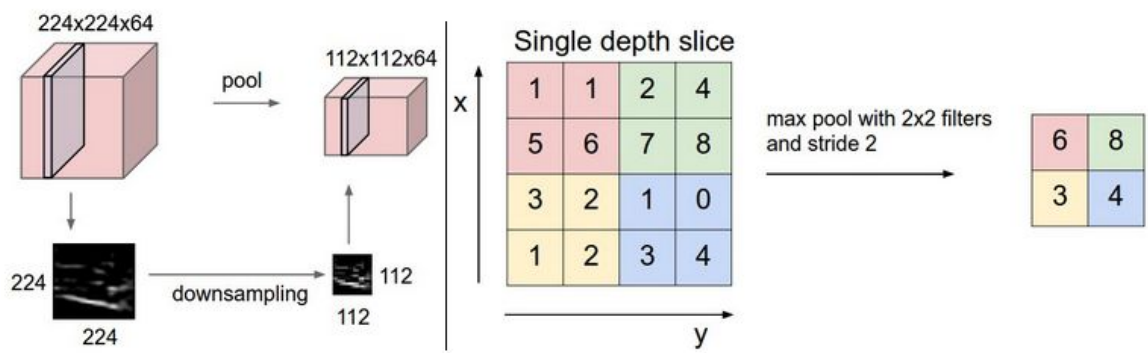

Fig. 55: Downsampling through pooling layer.
Source: http://cs231n.github.io/convolutional-networks/

Apart from downsampling, Pooling layer also helps in improving computational performance and prevents overfitting. Depth of the model is left invariant.

Following are the hyperparameters in pooling:
-   Stride
-   Pooling window
-   Type of pooling(Max vs Avg)

There are usually 4 types of poolings used:
●   Max Pooling
●   Average Pooling
●   L2 Norm of neighbourhood
●   Weighted avg. based on distance from central pixel

Pooling that pools over multiple features that are learned with several parameters can become rotational invariant as well. Rotational invariance is required in digit recognition etc.

### 4.2.1.Hardware implementation of pooling layers

**Max pooling layer:**
We used CMOS-JFET hybrid circuits to implement the function of max pooling layer with 9 inputs, thus can process a 3X3 block(pooling window) at a time. First stage of the circuit consists of a array of diodes with their outputs shorted. Usually, for fabricating diodes using CMOS technology, there are 3 different methods: nwell on P Substrate, p diffusion on Nwell and  n diffusion on P Substrate. Diodes fabricated using nwell on P Substrate have a significantly lower forward drop voltage and thus they can be used to implement the array of diode in this circuit. Next stage, consists of a voltage controlled current source. JFET gives a wider linear region for which input voltage is directly proportional to output current through the load R3.

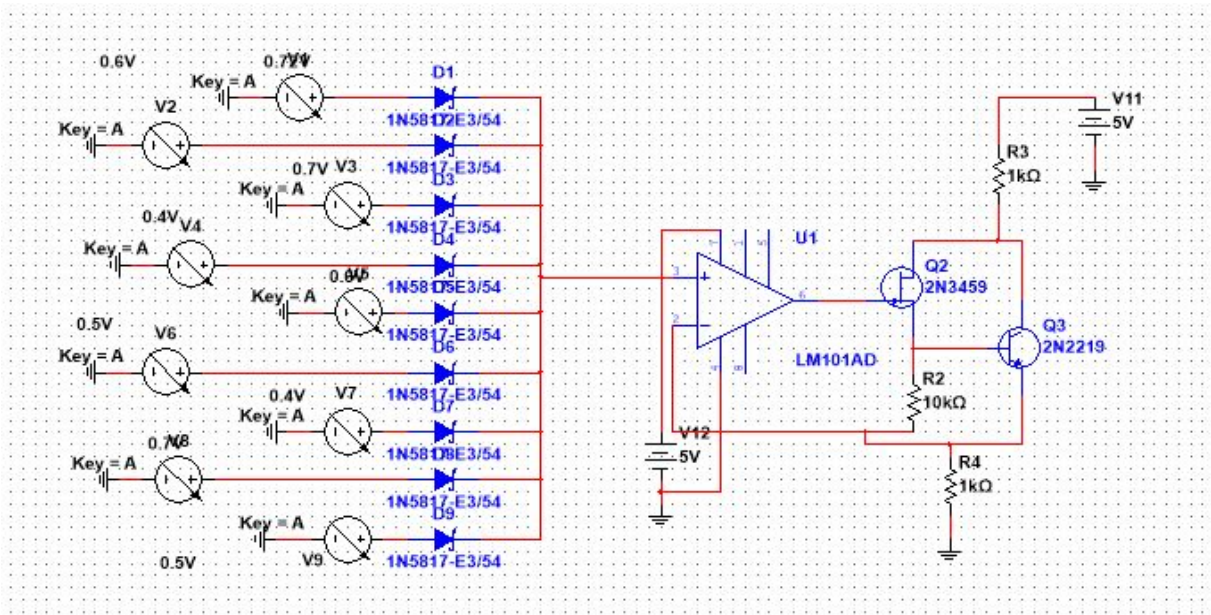

Fig 56: Max Pooling hardware implementation using CMOS and JFETs

Figure 57 shows the DC transfer characteristics of the above implementation of max pooling layer. DC characteristics were carried out by keeping all the voltages at 0.7V and then sweeping V1 from 0 to 2V. Graph shows that when V1 is below 0.7V, output current is constant and above that output current is linearly proportional to input voltage.

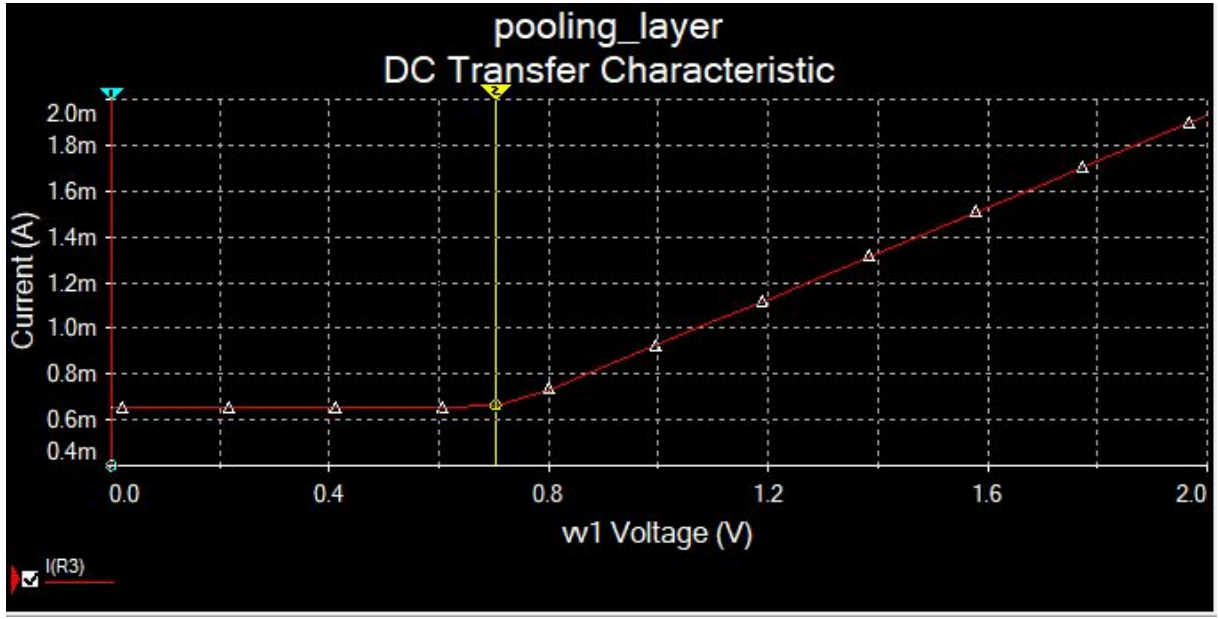

Fig. 57: DC characteristic of the max pooling layer circuit

**Average pooling layer:**

Average pooling layer can be implemented by using the 3X3 kernel sized convolutional filter with all the weights as 1/9. Circuit of convolutional filter is explained in the later chapters. Limitation of this circuit lies in the fact that with this circuit only 2D Local Average pooling layer with a pooling window of 3X3 or 2X2 can be implemented. Thus, implementation of Global and 3D pooling layers still remain an open problem.

**Normalization Layer:**

It is seen that distribution of each layer's input changes as learning parameters in the previous layers changes(called covariance shift), this causes non-uniformity and decreases learning rate. This problem can be solved by using a normalization layer. Normalization layer basically forces the layer's output to have unit standard deviation and zero mean. Normalization layers is used a pre-processing block to improve learning rates and reduce dependency on initialization.

**Dropout layer:**

Dropout in neural networks simply refers to dropping out several neurons during training by choosing dropout neuron at 'Random'.
In a fully connected NN, neurons develop inter-dependencies with each other which imparts redundancy, To prevent this, dropout is used. This prevents overfitting.

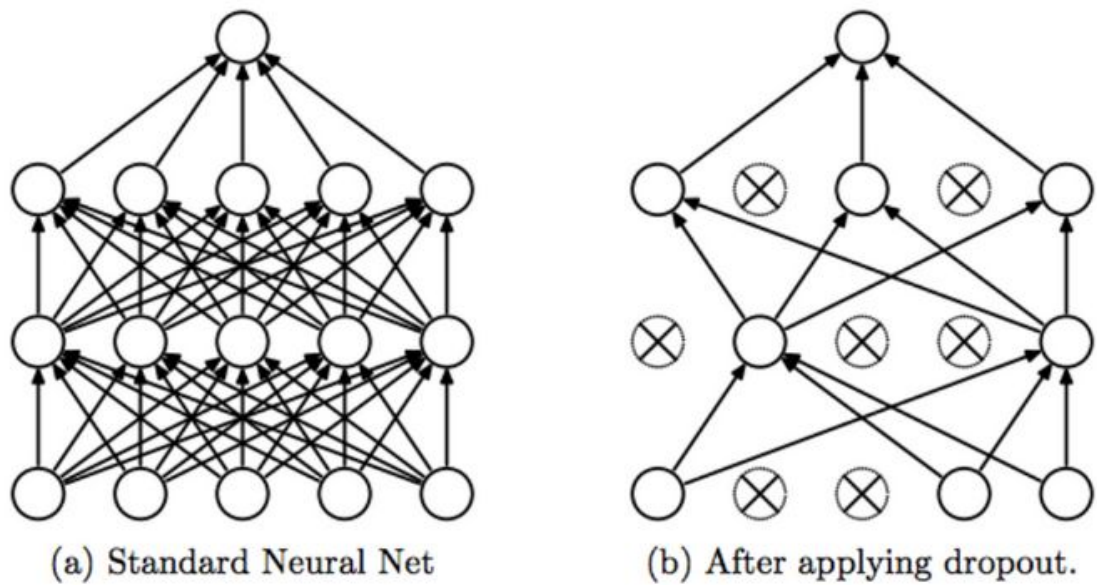

Fig. 58: Effect of Dropout on neuron interconnection
{Source: **https://tinyurl.com/y7j92vlu**}

Problem with dropout is that it nearly doubles the number of iterations required to converge.

# 5. Results and Conclusion

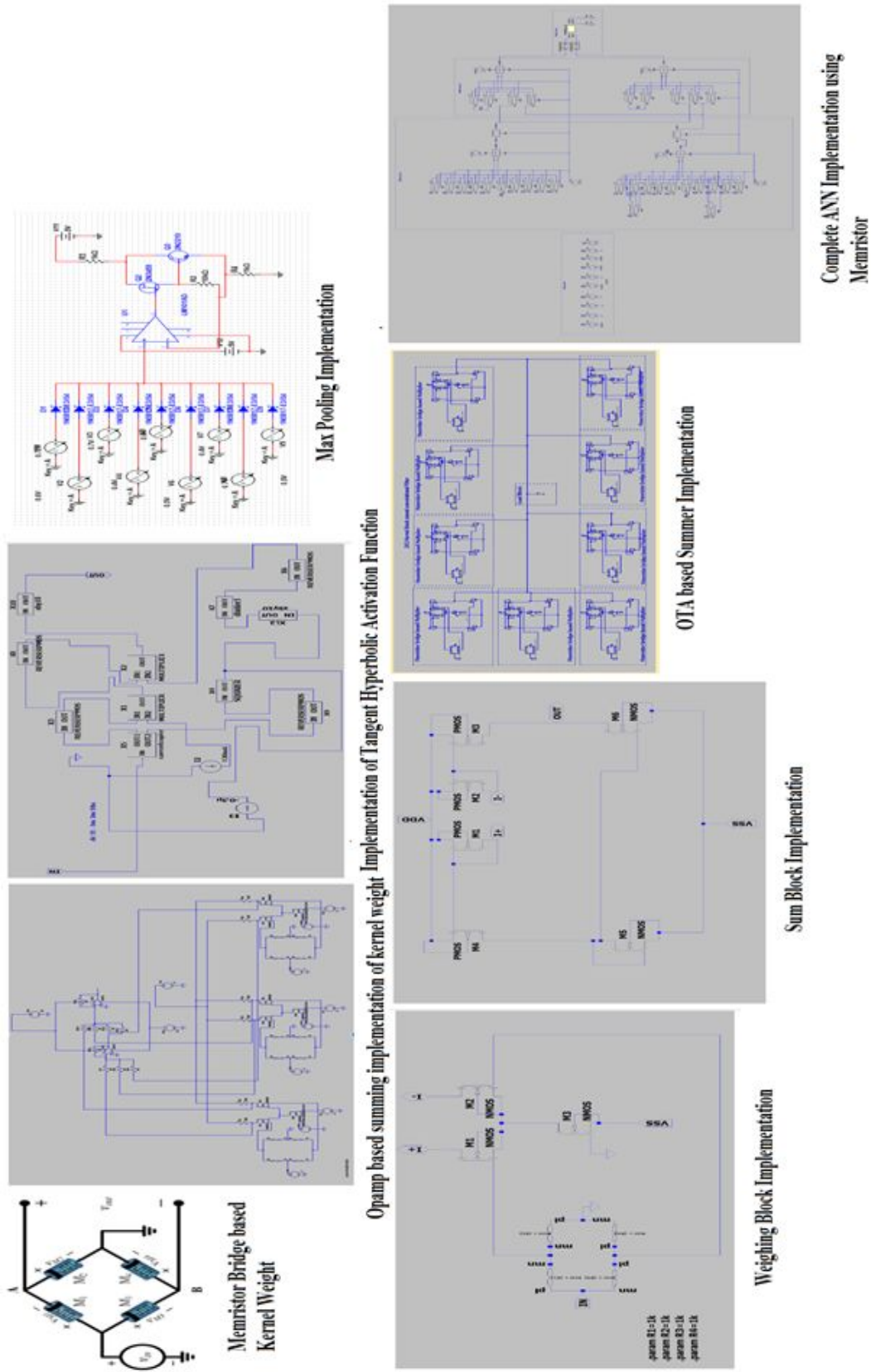

Figure 59: Compendium of different neuromorphic architectural elements proposed

The project revolves around constructing a memristor based model of a complete hardware implementation of a neural network accelerator. This project proposes basic building blocks based on preexisting memristor programming structures which have been utilised to create basic neural network primitives. After carrying out an extensive survey of state-of-the-art CMOS-Memristor based neural network and different building blocks, we proposed the implementation of a complete artificial neural network to classify tumours based on biopsy results. The circuit based neural network achieved an accuracy of 94.85% compared to the 96.71% accuracy of the original network. The trained weights were programmed onto the weight blocks of the memristor bridge in this neural network. The proposed method presented also expands the domain of the neuromorphic circuits by creating a complete hardware implementation of the multiple convolutional neural network layers that include image kernel based feature extraction, pooling and activation layers. Moreover, we have proposed two hardware topologies for integrating various existing multiplication block (memristor bridge neural network, proposed by Sah, 2012[]) for feature extraction using OTA and active load topologies. Moreover, this implementation was expanded to create 3x3 image kernels, for feature extraction. These features are then reduced using the pooling layer for reducing the data set size, for making practical computation possible. Moreover, the design of Rectified Linear Unit (ReLU) and Tangent Hyperbolic Circuit (Tanh) was made using hardware. Thus, this project proposes a complete hardware implementation of a CMOS-memristor based multipurpose configurable neural network architecture.

**Future Scope**
The further work possible in the proposed architecture includes creating memory buffers that can aid in storing the intermediary results of each stage of the neural network, due to the large computational time of each stage. Moreover, circuit reduction and optimization are important aspects of designing the neuromorphic circuit. Optimal handling of the non-linearities of the multiple stages must also be accounted for.

# Appendix

## A.1. Matlab Script for training the Artificial Neural Network:

```
1    % Solve a Pattern Recognition Problem with a Neural Network
2    % Script generated by Neural Pattern Recognition app
3    %
4    % This script assumes these variables are defined:
5    %
6    %   cancerInputs - input data.
7    %   cancerTargets - target data.
8
9    x = cancerInputs;
10   t = cancerTargets;
11
12   % Choose a Training Function
13   % For a list of all training functions type: help nntrain
14   % 'trainlm' is usually fastest.
15   % 'trainbr' takes longer but may be better for challenging problems.
16   % 'trainscg' uses less memory. Suitable in low memory situations.
17   trainFcn = 'trainscg';  % Scaled conjugate gradient backpropagation.
18
19   % Create a Pattern Recognition Network
20   hiddenLayerSize = 2;
21   net = patternnet(hiddenLayerSize);
22   net.layers{1}.transferFcn = 'poslin';
23   net.layers{2}.transferFcn = 'softmax';
24   % Setup Division of Data for Training, Validation, Testing
25   net.divideParam.trainRatio = 70/100;
26   net.divideParam.valRatio = 15/100;
27   net.divideParam.testRatio = 15/100;
28   % Train the Network
29   [net,tr] = train(net,x,t);
30   % Test the Network
31   y = net(x);
32   e = gsubtract(t,y);
33   performance = perform(net,t,y)
34   tind = vec2ind(t);
35   yind = vec2ind(y);
36   percentErrors = sum(tind ~= yind)/numel(tind);
37   % View the Network
38   view(net)
```

## A.2. Octave Script for Image Kernel Processing

## A.2.1. Image to PWL signals and software result of kernel

```
1  %Image to voltage signal (PWL)
2  clc;
3  clear;
4  voltage =5;
5  img = imread("testimg2.png");
6  gray_img = rgb2gray(img);
7  filtersize = [3 3];
8  imsize = size(gray_img);
9  oconv_size = [ (imsize(1) - filtersize(1) + 1) (imsize(2) - filtersize(2) + 1)];
10 %Extracting 3x3 segments from the image
11 for i = 1:1:oconv_size(1)
12    for j = 1:1:oconv_size(2)
13      for k = 1:1:filtersize(1)
14        for l = 1:1:filtersize(2)
15          imgseg(i,j,k,l) = gray_img( i+k-1, j+l-1);
16        end
17      end
18    end
19 end
20 %Signal Generation
21 fname="PWL";
22 fname1="";
23 imgseg1 = double(imgseg);
24 imgseg1 = (imgseg1 / 255)*voltage;
25 time = 0:0.001:((oconv_size(1)*oconv_size(2))/1000 - 0.001);
26 time = time';
27 for i= 1:1:filtersize(1)
28    for j=1:1:filtersize(2)
29      fname1 = strcat(fname,int2str(i),int2str(j));
30      tseq = [time reshape((imgseg1(:,:,i,j))',[],1)];
31      save('-ascii',fname1,'tseq');
32    end
33 end
34 %PC convolution :
35 %EDGE FILTER
36 imgseg = double(imgseg);
37 filter = [-1 -1 -1; -1 -8 1; -1 -1 -1];
38 filter = filter/10;
39 filter = reshape(filter,1,1,3,3);
40 for i=1:1:oconv_size(1)
41    for j=1:1:oconv_size(2)
42      %POSSIBLE RESHAPING ERROR:
43      conv_img(i,j) = sum(sum(imgseg(i,j,:,:).*filter));
44    end
45 end
46 conv_img = conv_img - min(min(conv_img));
47 conv_max = max(max(conv_img));
48 conv_img = conv_img * (255/conv_max);
49 conv_img = uint8(conv_img);
50 imshow(conv_img);
```

## A.2.2. PWL signals to image conversion

```
1   fname = "gridout.csv";
2   M = csvread(fname);
3   outdim = [477 477];
4   osize = outdim(1) * outdim(2);
5   seq1 = 0:0.001:(osize/1000 - 0.001);
6   in1 = interp1(M(:,1)',M(:,2)',seq1);
7   in1 = in1 - min(in1);
8   in1 = in1/max(in1);
9   in1 = in1*255;
10  in1 = uint8(in1);
11  img_out = reshape(in1,outdim(2),outdim(1));
12  img_out = img_out';
13  imshow(img_out);
```